\documentclass[10pt,twocolumn,letterpaper]{article}

\usepackage{cvpr}
\usepackage{times}
\usepackage{amssymb}
\usepackage{amsmath}
\usepackage{enumitem}
\usepackage{graphicx}
\usepackage{caption}
\usepackage{subcaption}
\usepackage{textcomp}
\usepackage[activate]{pdfcprot}
\usepackage{fixltx2e}
\usepackage{tabularx}
\usepackage{bm}
\newcolumntype{P}[1]{>{\raggedright\arraybackslash}m{#1}}%
\newcolumntype{C}[1]{>{\centering\arraybackslash}m{#1}}%
\newcolumntype{R}[1]{>{\raggedleft\arraybackslash}m{#1}}%

\usepackage[pagebackref=true,breaklinks=true,colorlinks,bookmarks=false]{hyperref}

\newcommand{\owntitle}{Appearance-Based Gaze Estimation in the Wild}

\newcommand{\ownkeywords}{Appearance-based gaze estimation, daily life settings, multimodal convolutional neural networks}
\newcommand{\numberpictures}{213,659}

\newcommand{\numbermannualannotation}{10,848}
\newcommand{\datasetname}{MPIIGaze}

\newcommand{\thickhline}{%
    \noalign {\ifnum 0=`}\fi \hrule height 1pt
    \futurelet \reserved@a \@xhline
}
\newcolumntype{"}{@{\hskip\tabcolsep\vrule width 2pt\hskip\tabcolsep}}

\cvprfinalcopy

\ifcvprfinal\fi

\hypersetup{%
pdftitle={\owntitle},
pdfauthor={Xucong Zhang, Yusuke Sugano, Mario Fritz, Andreas Bulling},
pdfkeywords={\ownkeywords},
bookmarksnumbered,
pdfstartview={FitH},
colorlinks,
citecolor=red,
filecolor=red,
linkcolor=red,
urlcolor=red,
breaklinks=true
}

\begin{document}

\title{\owntitle}

\author{
Xucong Zhang$^1$
\and
Yusuke Sugano$^1$
\and
Mario Fritz$^2$
\and
Andreas Bulling$^1$\\
\and
$^1$Perceptual User Interfaces Group, $^2$Scalable Learning and Perception Group\\
Max Planck Institute for Informatics, Saarbr\"ucken, Germany\\
{\tt\small \{xczhang,sugano,mfritz,bulling\}@mpi-inf.mpg.de}
}

\maketitle

\begin{abstract}
Appearance-based gaze estimation is believed to work well in real-world settings, but existing datasets have been collected under controlled laboratory conditions and methods have been not evaluated across multiple datasets. In this work we study appearance-based gaze estimation in the wild. We present the \datasetname~dataset that contains \numberpictures~images we collected from 15 participants during natural everyday laptop use over more than three months. Our dataset is significantly more variable than existing ones with respect to appearance and illumination. We also present a method for in-the-wild appearance-based gaze estimation using multimodal convolutional neural networks that significantly outperforms state-of-the art methods in the most challenging cross-dataset evaluation. We present an extensive evaluation of several state-of-the-art image-based gaze estimation algorithms on three current datasets, including our own. This evaluation provides clear insights and allows us to identify key research challenges of gaze estimation in the wild.
\end{abstract}

\section{Introduction}

Appearance-based gaze estimation is well established as a research topic in computer vision because of its relevance for several application domains, including gaze-based human-computer interaction and visual behaviour analysis~\cite{morimoto2005eye}.
Purely learning-based methods were recently proposed to learn generic gaze estimators from large amounts of person, and head pose-independent training data~\cite{odobez2013person,schneider2014manifold,suganolearning}.
Such methods have the potential to bring appearance-based methods into settings that do not require any user- or device-specific training.
Gaze estimation using monocular cameras is particularly promising given the proliferation of such cameras in hand-held and portable devices, such as mobile phones and laptops, as well as interactive displays.

\begin{figure}[t]
\center
\includegraphics[width=\columnwidth]{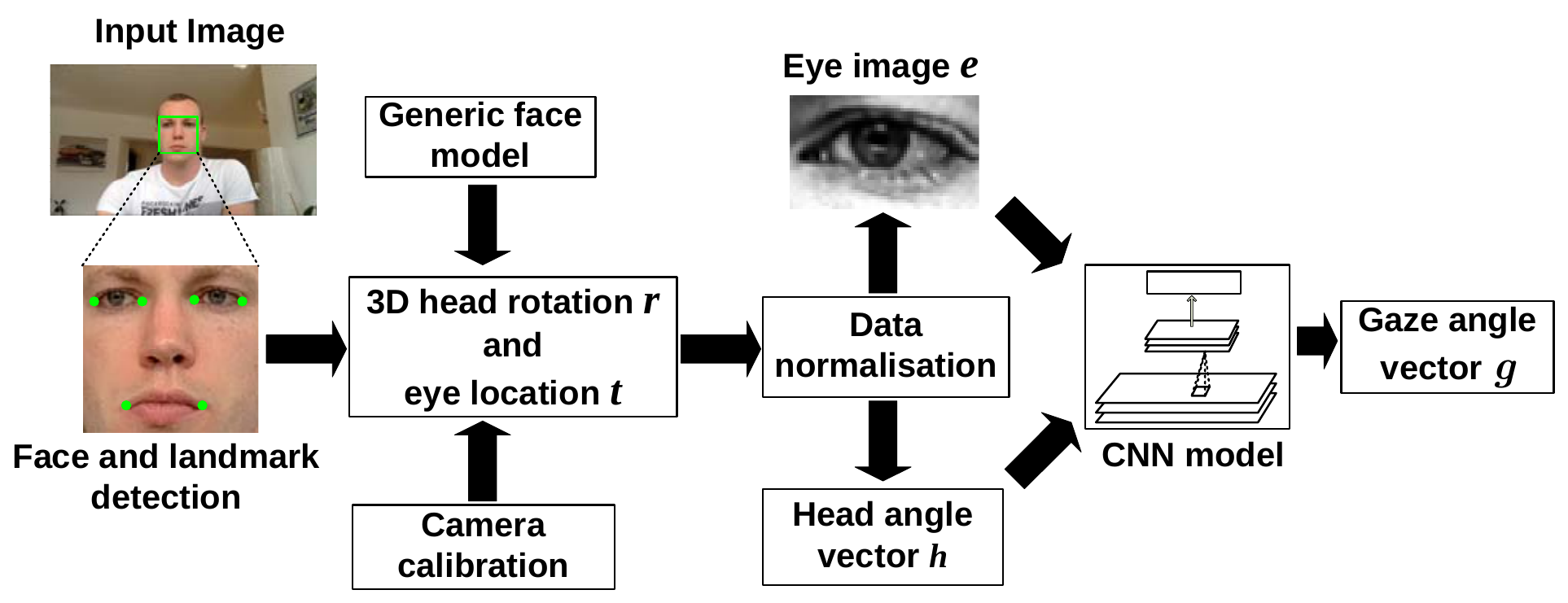}
\caption{Overview of our method for in-the-wild appearance-based gaze estimation using multimodal convolutional neural networks.}
\label{fig:pipeline}
\end{figure}

While appearance-based gaze estimation is believed to perform well in everyday settings, state-of-the-art learning-based methods are still developed and evaluated on datasets collected under controlled laboratory conditions. These conditions are characterised by limited variability of eye appearances as well as the assumption of accurate head pose estimates.
Current appearance-based gaze estimation methods are also not evaluated across different datasets, which bears the risk of significant dataset bias -- a key problem
also in object recognition~\cite{torralba2011unbiased} and salient object detection~\cite{li2014secrets}.

\renewcommand{\tablename}{Figure}
\setcounter{table}{1}
\setcounter{figure}{2}
\begin{table*}[ht]
\hskip-0.25cm\begin{tabularx}{\textwidth}{C{2.6cm} C{2.6cm} C{2.6cm} C{2.6cm} C{2.6cm} " C{2.6cm}}
\includegraphics[width=2.6cm]{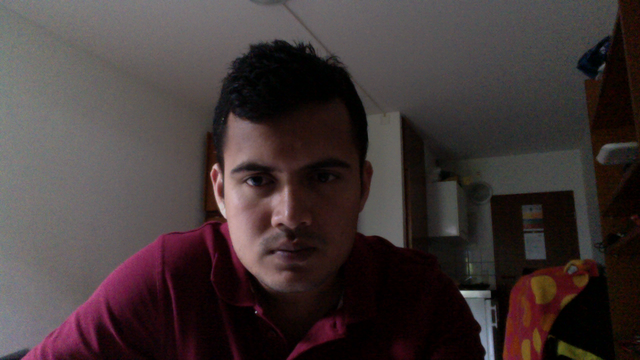} &
\includegraphics[width=2.6cm]{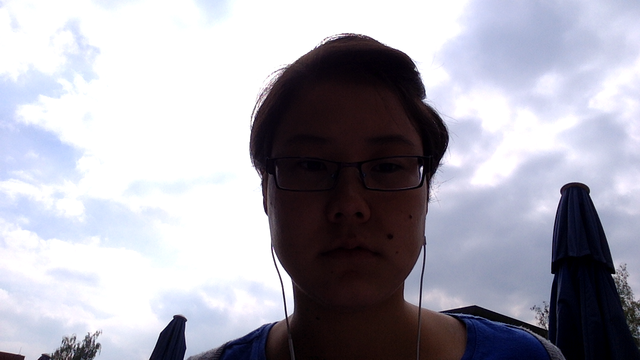} &
\includegraphics[width=2.6cm]{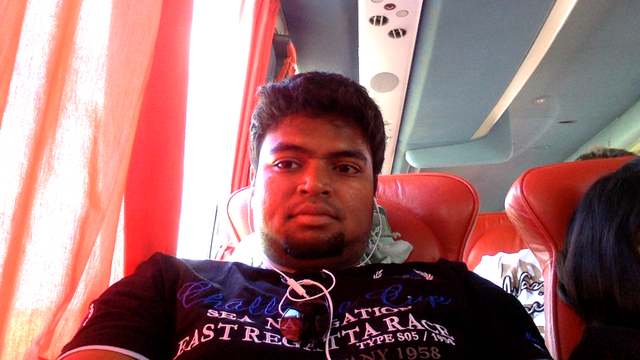} &
\includegraphics[width=2.6cm]{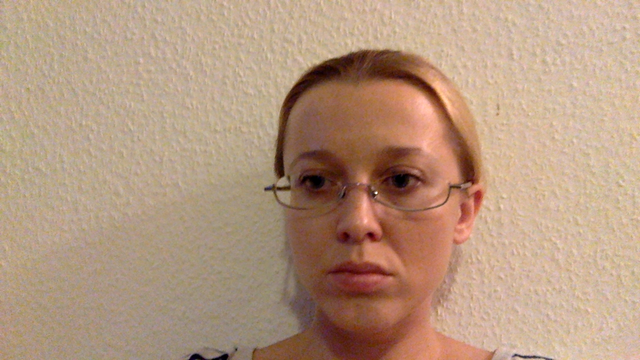} &
\includegraphics[width=2.6cm]{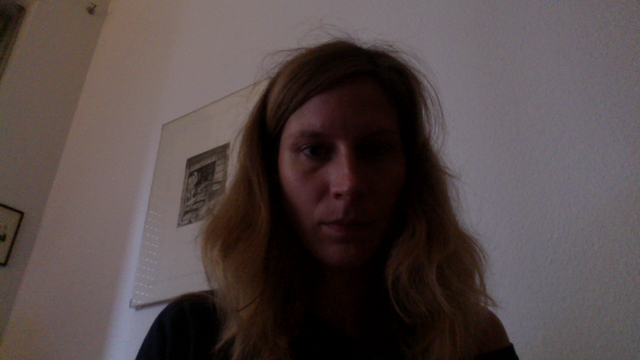} &
\includegraphics[width=2.6cm]{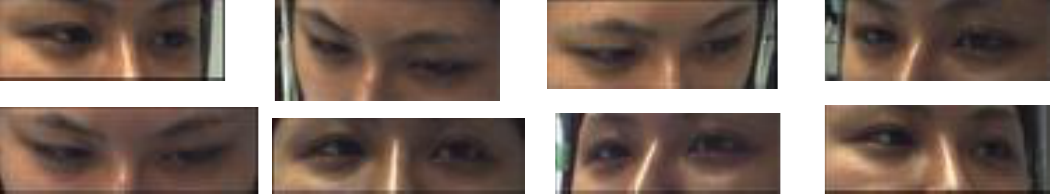} \\
\includegraphics[width=2.6cm]{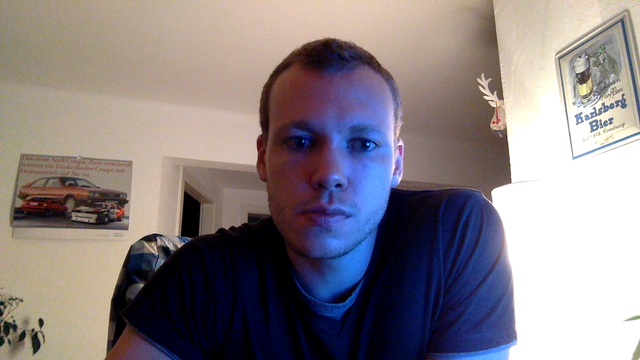} &
\includegraphics[width=2.6cm]{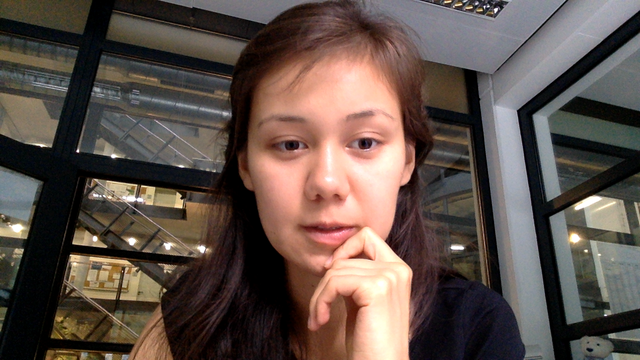} &
\includegraphics[width=2.6cm]{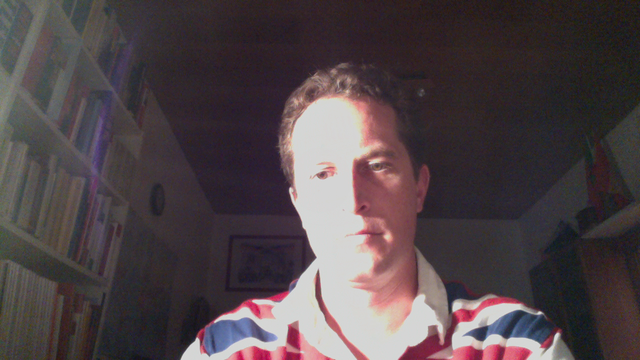} &
\includegraphics[width=2.6cm]{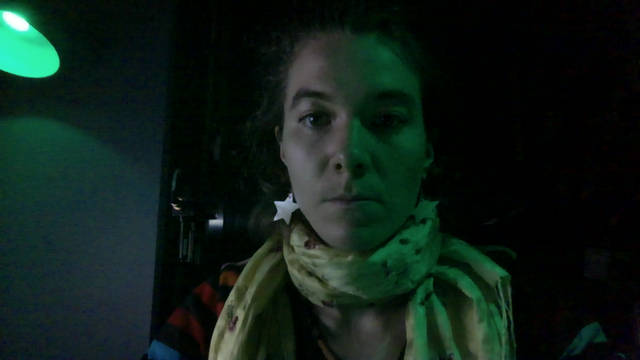} &
\includegraphics[width=2.6cm]{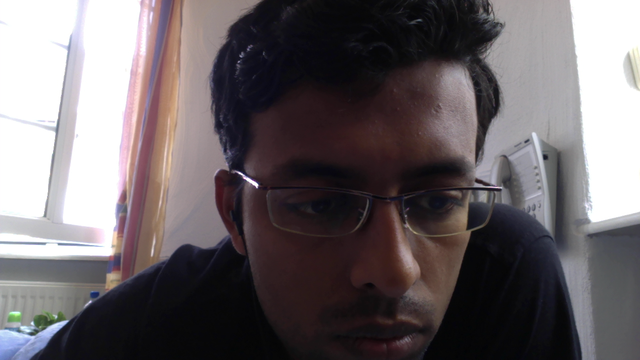} &
\includegraphics[height=1.5cm]{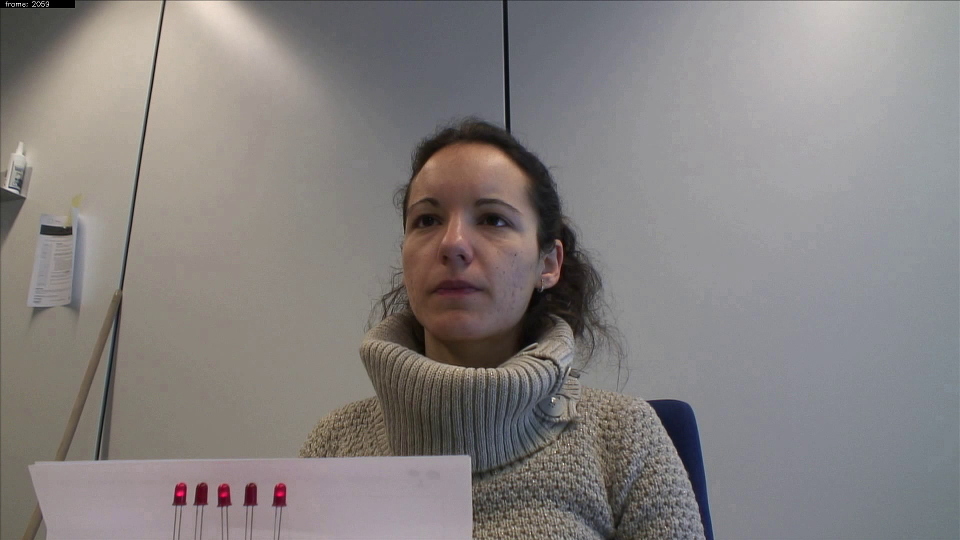} \\
\includegraphics[width=2.6cm]{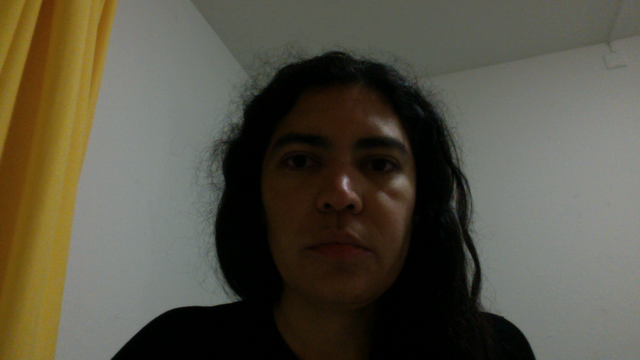} &
\includegraphics[width=2.6cm]{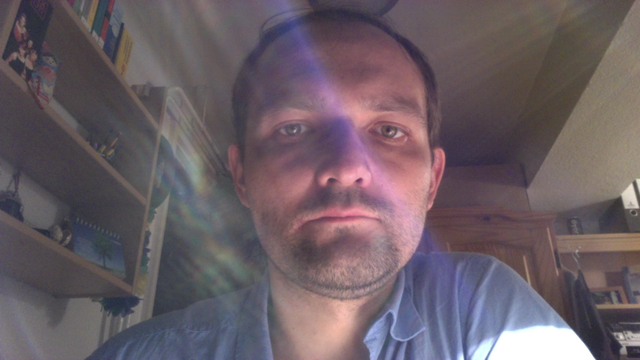} &
\includegraphics[width=2.6cm]{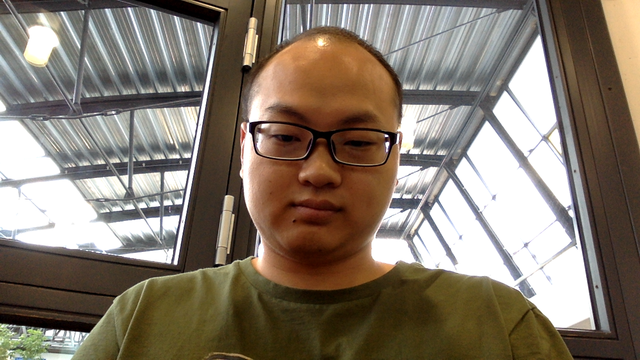} &
\includegraphics[width=2.6cm]{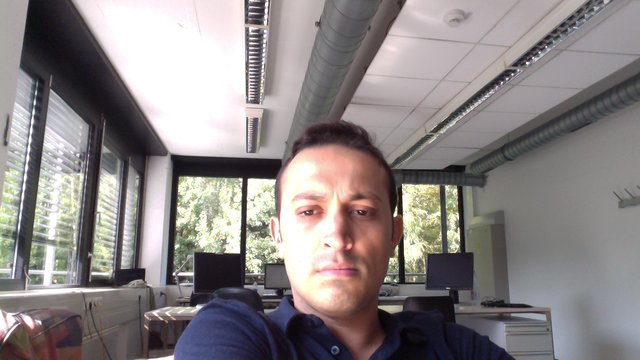} &
\includegraphics[width=2.6cm]{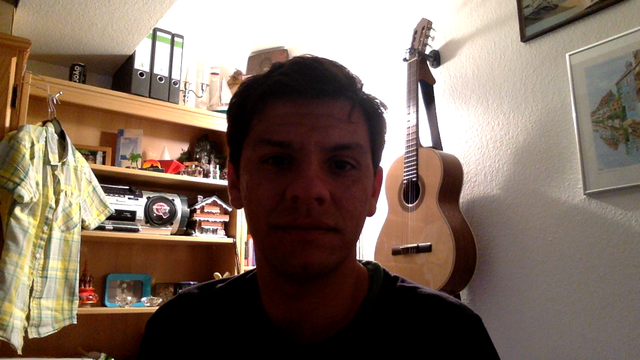} &
\includegraphics[height=1.5cm]{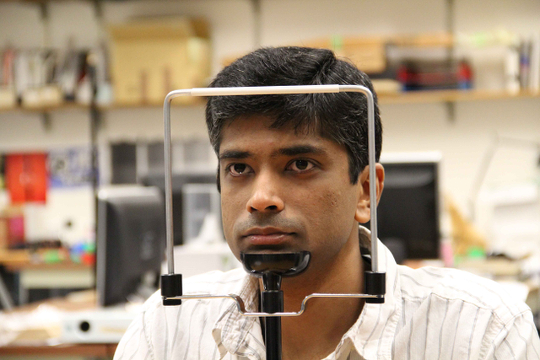} \\
\end{tabularx}
\caption{Sample images from our \datasetname~dataset showing the considerable variability in terms of place and time of recording, directional light and shadows. For comparison, the last column shows sample images from other current publicly available datasets (cf. Table~\ref{tab:other_datasets}): UT Multiview~\cite{suganolearning} (top), Eyediap~\cite{FunesMora_ETRA_2014} (middle), Smith et al.~\cite{smith2013gaze} (bottom).}
\label{fig:teaser}
\end{table*}
\renewcommand{\tablename}{Table}
\setcounter{table}{0}

In this work we make the first step towards appearance-based gaze estimation in the wild. Given a lack of realistic data, we created the \datasetname~dataset that contains \numberpictures~images collected from 15 laptop users over several months (see Figure~\ref{fig:teaser}). \datasetname~covers a realistic variability in appearance and illumination and therefore represents a significant advance over existing datasets. Laptops not only allow us to record in the wild but they also have considerable potential as an application platform, such as for gaze interaction~\cite{majaranta2014eye}.
The dataset and annotations are publicly available online.

We study two key tasks through extensive evaluations of appearance-based gaze estimation algorithms on three publicly available gaze estimation datasets:

\begin{enumerate}[topsep=0.2cm,parsep=0.2cm,partopsep=0.2cm,itemsep=0.2cm,labelindent=0pt]
\item Handling appearance differences between training and testing data. Since we cannot always assume a training dataset that can cover the whole test space, the important question is how robustly the estimator can handle unknown appearance conditions.
\item Pursuing the performance gain for domain-specific training. If we can assume that training data is directly collected in the target daily-life environment, the goal is to fully utilise the rich training data.
\end{enumerate}

While better performances can be expected for the second domain-specific training task where both the training and testing data come from the same dataset, the ultimate goal of person-independent gaze estimation is to handle the first cross-domain training task, which leads to the most challenging but practically most important use cases.

The contribution of this work is threefold. First, we introduce the first large-scale dataset for appearance-based gaze estimation in the wild. Our dataset is one order of magnitude larger than existing datasets and significantly more variable with respect to illumination and appearance.
Second, we present an extensive evaluation of state-of-the-art gaze estimation algorithms on three current datasets, including our own, and identify key research challenges of in-the-wild settings. 
Third, we present a method for appearance-based gaze estimation that uses multimodal convolutional neural networks and that significantly outperforms state-of-the-art methods in the most challenging cross-dataset evaluation.

\section{Related Work}

\begin{table*}[ht]
\footnotesize
\centering
\begin{tabularx}{\textwidth}{P{3.5cm} C{2cm} C{2.4cm} C{1.8cm} C{1.8cm} C{1.4cm} R{1.2cm}}
& \textbf{Participants} & \textbf{Head poses} & \textbf{On-screen gaze targets} & \textbf{Illumination conditions} & \textbf{Duration (days)} & \textbf{Images}\\
\hline
McMurrough et al.~\cite{McMurrough:2012:ETD:2168556.2168622} & 20 & 1 & 16 & 1 & 1 & videos\\
\hline
Villaneuva et al.~\cite{villanueva2013hybrid} & 103 & 1 & 12 & 1 & 1 & 1,236\\
\hline
Weidenbacher et al.~\cite{weidenbacher07} & 20 & 19 & 2-9 & 1 & 1 & 2,220\\
\hline
Smith et al.~\cite{smith2013gaze} & 56 & 5 & 21 & 1 & 1 & 5,880\\
\hline
Eyediap~\cite{FunesMora_ETRA_2014} & 16 & continuous & continuous & 2 & 1 & videos\\
\hline
UT Multiview~\cite{suganolearning} & 50 & 8 + synthesised & 160 & 1 & 1 & 64,000\\
\hline
\datasetname~(ours) & 15 & continuous & continuous & daily life & 45.7 & \numberpictures\\
\end{tabularx}
\caption{Comparison of current publicly available appearance-based gaze estimation datasets with respect to number of participants, head poses and on-screen gaze targets (discrete or continuous), number of different illumination conditions,  average duration of data collection per participant, and total number of images.}
\label{tab:other_datasets}
\end{table*}

\subsection{Gaze Estimation Methods}

Gaze estimation methods can be model-based or appearance-based~\cite{hansen2010eye}.
Model-based methods use a geometric eye model and can be further divided into corneal-reflection and shape-based methods, depending on whether they require external light sources to detect eye features. Early works on corneal reflection-based methods focused on stationary settings~\cite{shih2004novel,morimoto2002detecting,hennessey2006single,yoo2005novel} and were later extended to handle arbitrary head poses using multiple light sources or cameras~\cite{zhu2005eye,zhu2006nonlinear}. In contrast, shape-based methods~\cite{Ishikawa_2004_4705,Chen2008,yamazoe2008remote,valenti2012combining} directly infer gaze directions from observed eye shapes, such as pupil centre or iris edges.
Although they have recently been applied to more practical application scenarios~\cite{jianfeng2014eye,mora2014geometric,sun2014realtime,wood14_etra}, their accuracy is lower and it is unclear whether shape-based approaches can robustly handle low image quality and variable lighting conditions.
Appearance-based gaze estimation methods directly use eye images as input
and can therefore potentially work with low-resolution eye images.
While early works assumed a fixed head pose~\cite{baluja1994non,tan2002appearance,williams2006sparse,sewell10_chi,lu2014alr,Liang:2013:AGT:2509315.2509318}, recent works focused on methods for 3D head pose estimation~\cite{lu2014learning,lu2012head,funes2012gaze,choi2013appearance}. However, appearance-based methods require larger amounts of user-specific training data than model-based methods, and it remains unclear if the learned estimator can generalise to unknown users.
Similarly, previous methods typically assumed accurate 3D head poses as input, which is a strong assumption for unconstrained in-the-wild settings.

\subsection{Calibration-Free Gaze Estimation}

The requirement to collect person-specific training data during a calibration step is a key limitation of both model-based and appearance-based methods. To address this limitation, several previous works used interaction events, such as mouse clicks or key presses, as a proxy for the user's on-screen gaze position~\cite{sugano2008incremental,Huang:2014:BSE:2647868.2655031}. Alternatively, visual saliency maps~\cite{chen2011probabilistic,sugano2013appearance} or pre-recorded human gaze patterns of the presented visual stimuli~\cite{alnajarcalibration} were used as bottom-up, probabilistic training data to learn the estimation function. However, all of these approaches rely on observations of a specific person and environment, which limits their applicability.

Purely data-driven approaches leverage large amounts of training data to learn gaze estimators that generalise to arbitrary users without the need for person-specific calibration~\cite{schneider2014manifold,odobez2013person,suganolearning} settings. These methods have significant potential to bring gaze estimation to new settings, including mobile devices, public displays, and egocentric cameras. However, the generalization capability of learning-based methods has not been examined yet.
Moreover, prior work used 3D input for head pose information~\cite{odobez2013person,suganolearning}, while we are the first to evaluate the whole pipeline for fully automatic monocular appearance-based gaze estimation for person-independent training scenario.

\subsection{Datasets}

Because most existing gaze estimation datasets are designed for coarse gaze estimation, the sampling density of gaze and head pose space is not sufficient to train appearance-based gaze estimators~\cite{McMurrough:2012:ETD:2168556.2168622,villanueva2013hybrid,weidenbacher07,smith2013gaze} (see Table~\ref{tab:other_datasets} for an overview of existing datasets). More comparable to \datasetname, the Eyediap dataset contains 94 video sequences of 16 participants looking at three different targets (discrete and continuous markers displayed on a monitor, and floating physical targets) under both static and free head motion~\cite{FunesMora_ETRA_2014}. The UT Multiview dataset also contains dense gaze samples of 50 participants as well as 3D reconstructions of eye regions that can be used to synthesise images for arbitrary head poses~\cite{suganolearning}. However, as discussed before, both datasets have the significant limitation that they were recorded under controlled laboratory settings. Although the Eyediap dataset includes two different illumination conditions, recordings under the second condition were provided only for a subset of the participants.

\section{The \datasetname~dataset}

\begin{figure*}[t]
        \centering
        \begin{subfigure}[b]{0.33\textwidth}
                \includegraphics[width=\textwidth]{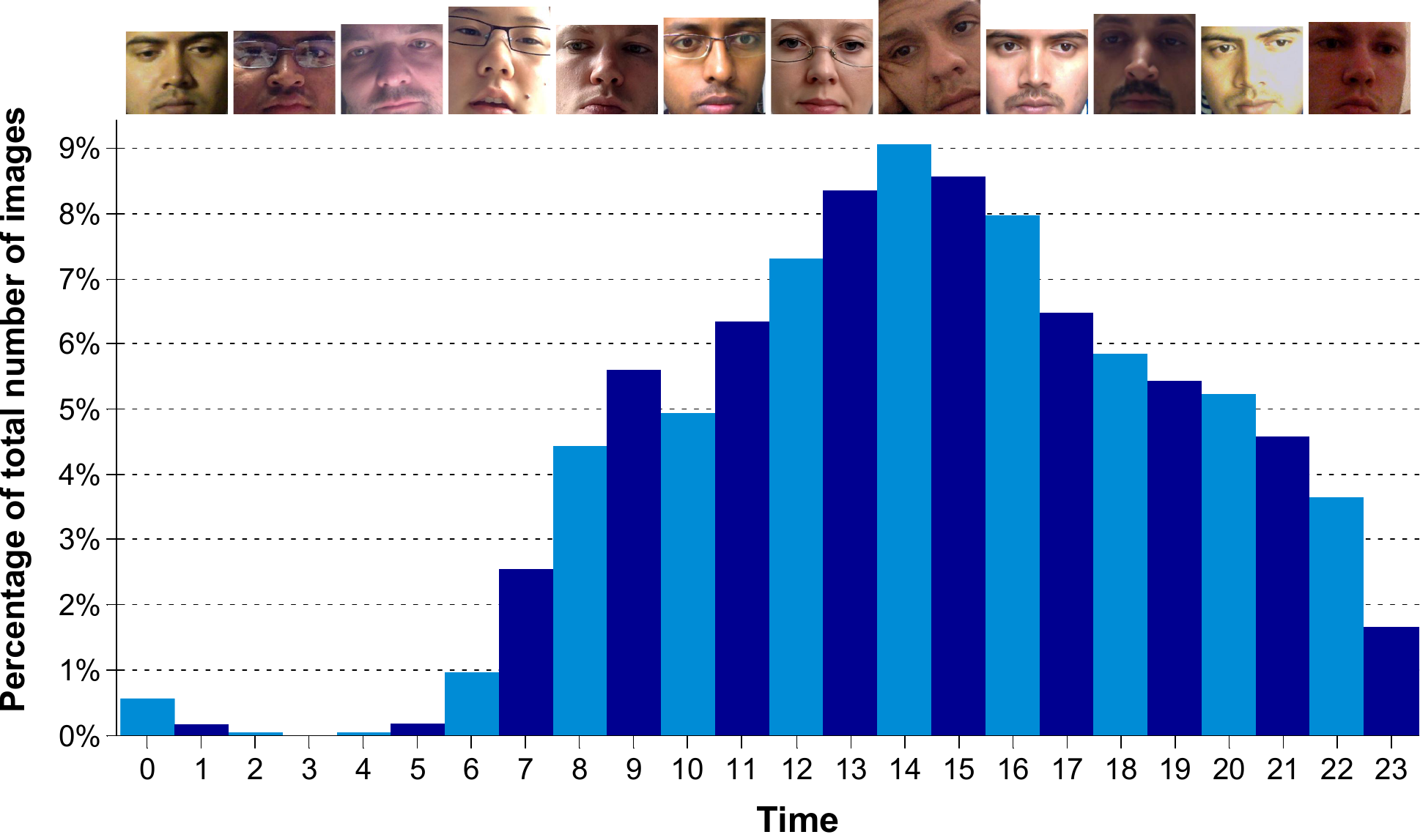}
                \label{fig:characteristics:timeofday}
        \end{subfigure}
        \begin{subfigure}[b]{0.33\textwidth}
                \includegraphics[width=\textwidth]{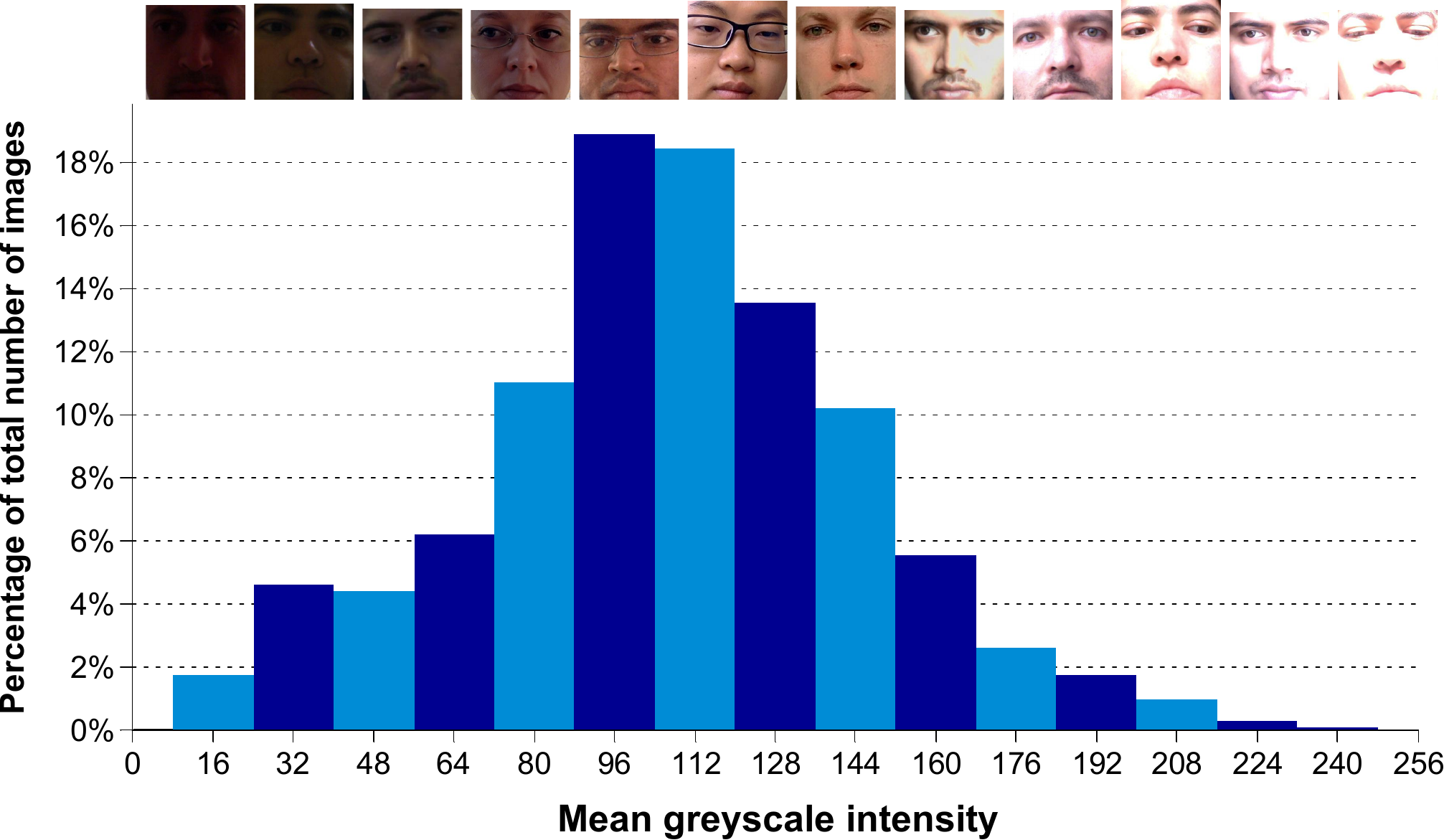}
                \label{fig:characteristics:intensity}
        \end{subfigure}
        \begin{subfigure}[b]{0.33\textwidth}
                \includegraphics[width=\textwidth]{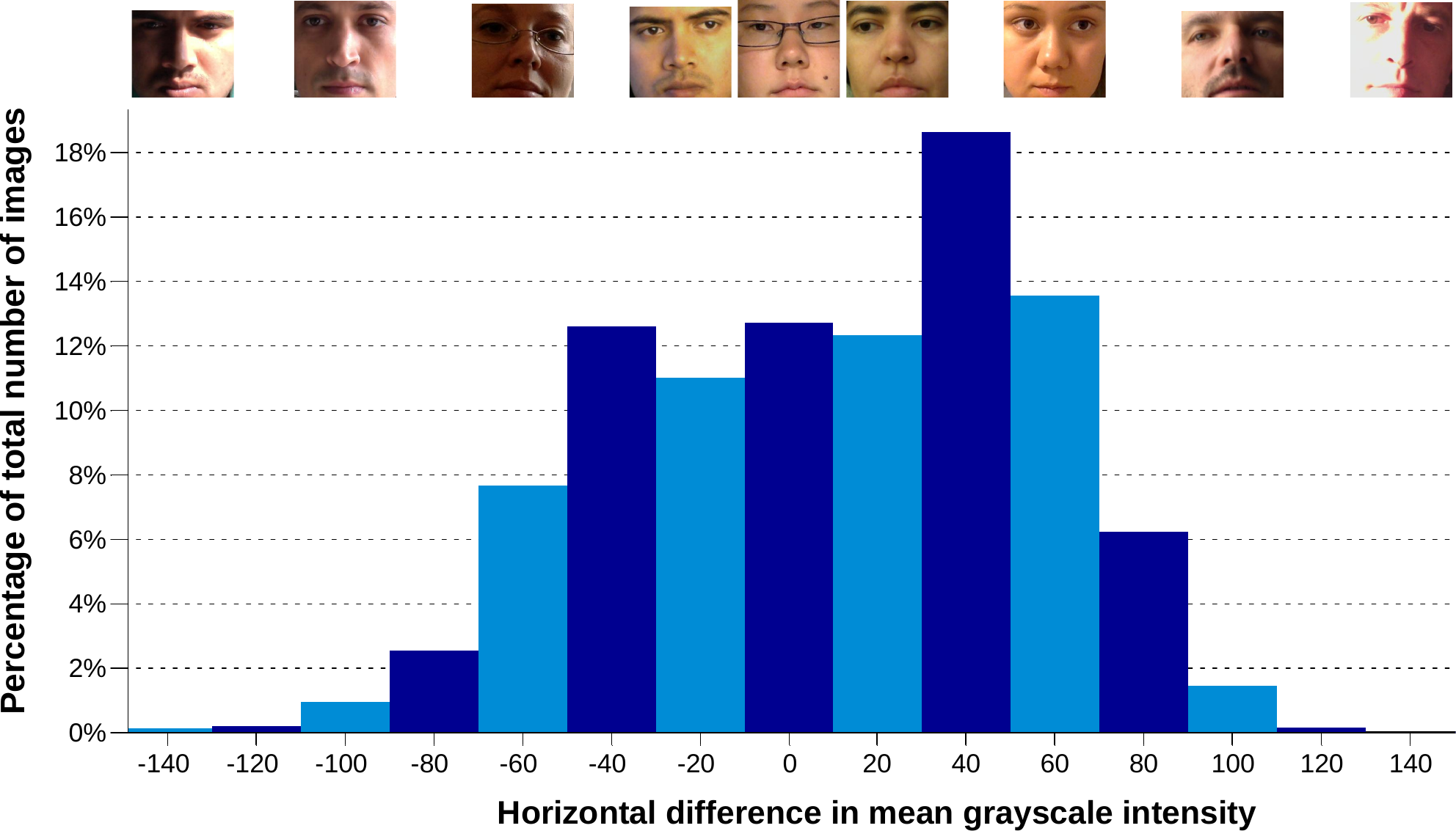}
                \label{fig:characteristics:directional}
        \end{subfigure}
        \caption{Key characteristics of our dataset. Percentage of images collected at different times over the day (left), having different mean grey-scale intensities within the face region (middle), and having horizontally different mean grey-scale intensities from the left to right half of the face region (right). Representative samples at the top.}
    \label{fig:characteristics}
\end{figure*}

We designed our data collection procedure with two main objectives in mind: 1) to record images of participants outside of controlled laboratory conditions, i.e during their daily routine, and 2) to record participants over several months to cover a wider range of recording locations and times, illuminations, and eye appearances. We opted for recording images on laptops not only because they are suited for long-term daily recordings but also because they are an important platform for eye tracking applications~\cite{majaranta2014eye}. Laptops are personal devices, therefore typically remaining with a single user, and they are used throughout the day and over long periods of time. They also come with high resolution front-facing cameras that are in a fixed position relative to the screen. We further opted to use an experience sampling approach to ensure images were collected regularly throughout the data collection period~\cite{larson1983experience}.

\subsection{Collection Procedure}

We implemented custom software running as a background service on participants' laptops. Every 10 minutes the software automatically asked participants to look at a random sequence of 20 on-screen positions (a recording session), visualised as a grey circle shrinking in size and with a white dot in the middle. Participants were asked to fixate on these dots and confirm each by pressing the spacebar once the circle was about to disappear. This was to ensure participants concentrated on the task and fixated exactly at the intended on-screen positions. 
No other instructions were given to them, in particular no constraints as to how and where to use their laptops. Because our dataset covers different laptop models with varying screen size and resolution, on-screen gaze positions were  converted to physical 3D positions in a camera coordinate system. We obtained the intrinsic parameters from each camera beforehand. 3D positions of each screen plane were estimated using a mirror-based calibration method~\cite{Rodrigues2010Mirror}.

We also asked human annotators to provide face annotations for a random subset of \numbermannualannotation~images to increase the value of the dataset for other tasks, such as face detection and alignment. They annotated these images with a total of 12 facial landmarks, following an extended LFW style~\cite{huang2008labeled}, that additionally contained a face bounding box and two eye bounding boxes, as well as the left and right pupil position.

\subsection{Dataset Characteristics}

We collected a total of \numberpictures~images from 15 participants. The number of images collected by each participant varied from 34,745 to 1,498.
Figure~\ref{fig:characteristics} (left) shows a histogram of times of the recording sessions. Although there is a certain bias towards working hours, the figure shows the high variation in recording times. Consequently, our dataset also contains larger variability in illumination. To visualise the different illumination conditions, Figure~\ref{fig:characteristics} (bottom) shows a histogram of mean grey-scale intensities inside the face region. Figure~\ref{fig:characteristics} (right) further shows a histogram of the mean intensity differences from the right side to the left side of the face region, which approximates the statistics of directional light sources. These figures underline the complexity of our dataset in terms of appearance variations.

\begin{figure}[t]
  \centering
        \begin{subfigure}[b]{0.33\linewidth}
                \includegraphics[width=\textwidth]{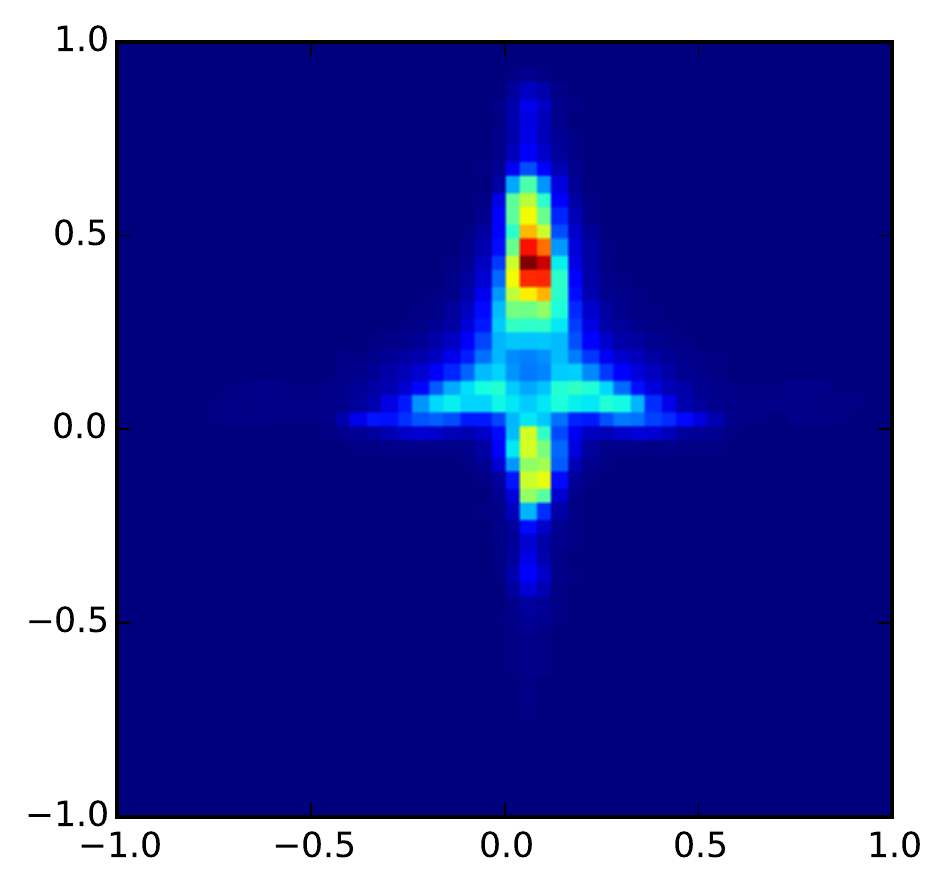}
                \caption{$\bm{h}$ (\datasetname)}
                \label{fig:dataset_comparison:mpii-pose}
        \end{subfigure}%
        \begin{subfigure}[b]{0.33\linewidth}
                \includegraphics[width=\textwidth]{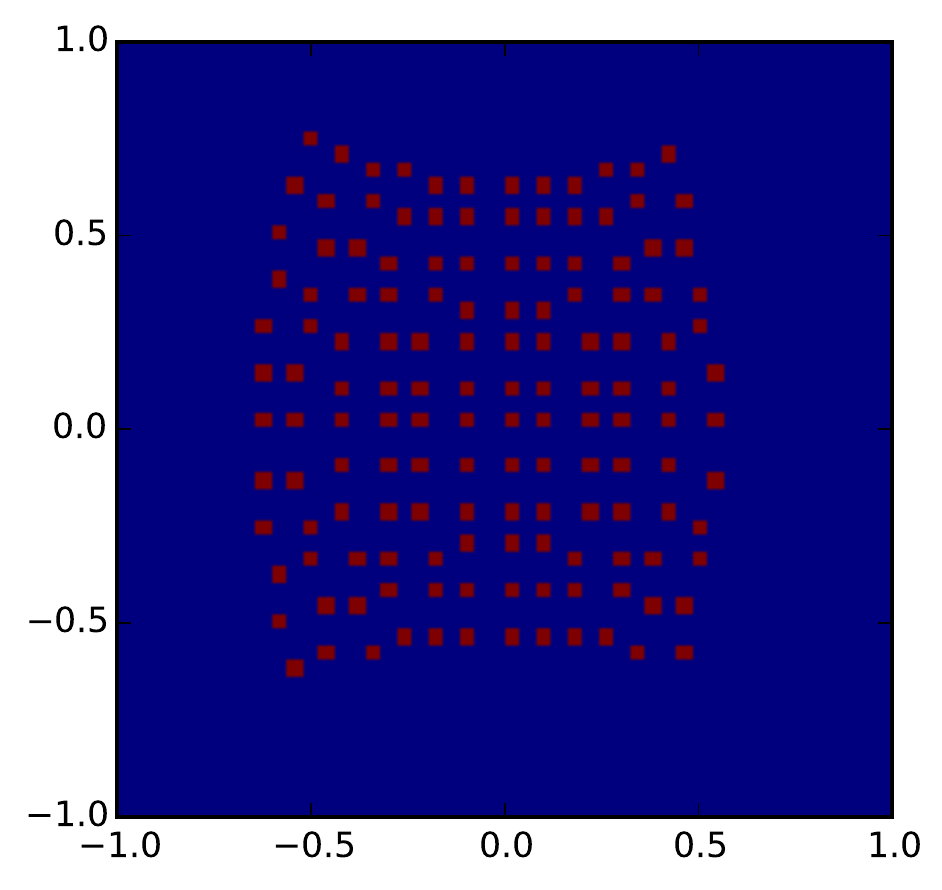}
                \caption{$\bm{h}$ (UT Multiview)}
                \label{fig:dataset_comparison:utfull-pose}
        \end{subfigure}%
        \begin{subfigure}[b]{0.33\linewidth}
                \includegraphics[width=\textwidth]{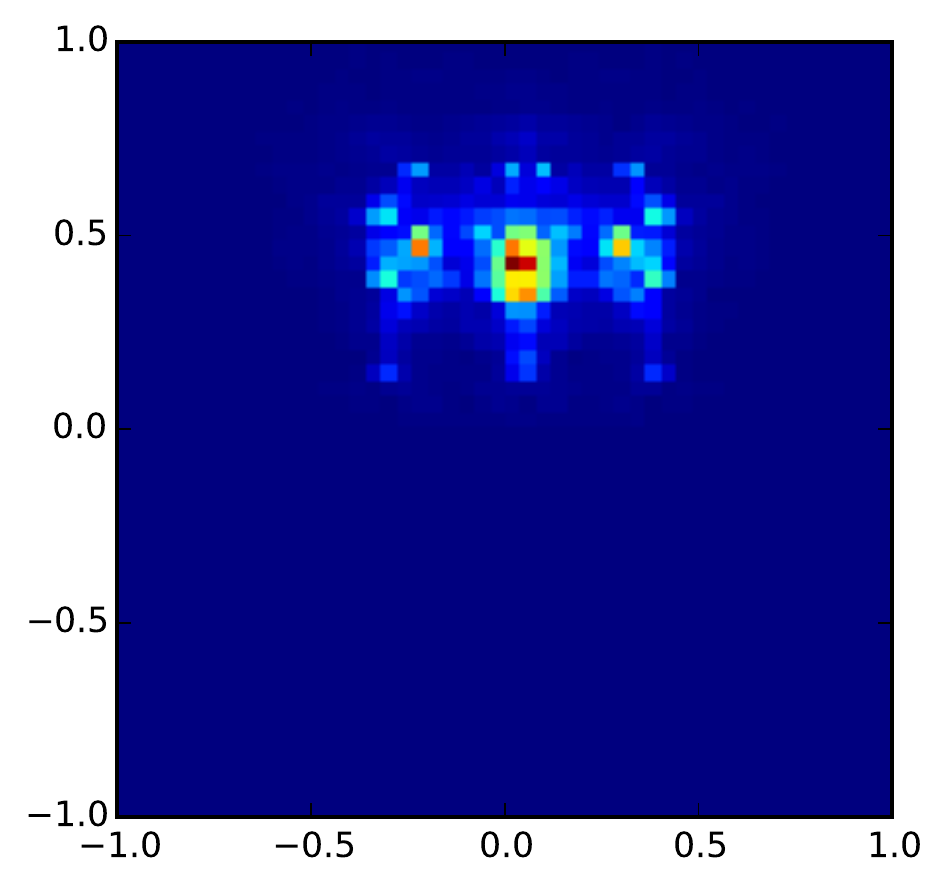}
                \caption{$\bm{h}$ (Eyediap)}
                \label{fig:dataset_comparison:idiap-ds-pose}
        \end{subfigure}%

        \begin{subfigure}[b]{0.33\linewidth}
                \includegraphics[width=\textwidth]{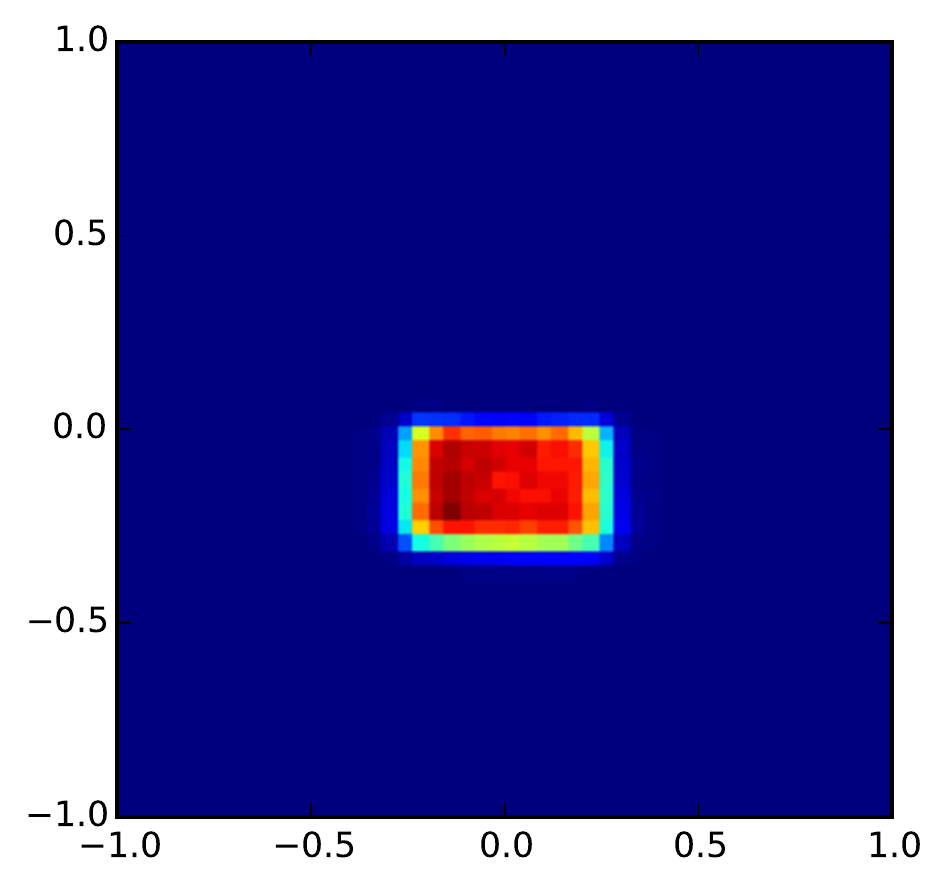}
                \caption{$\bm{g}$ (\datasetname)}
                \label{fig:dataset_comparison:mpii-gaze}
        \end{subfigure}%
        \begin{subfigure}[b]{0.33\linewidth}
                \includegraphics[width=\textwidth]{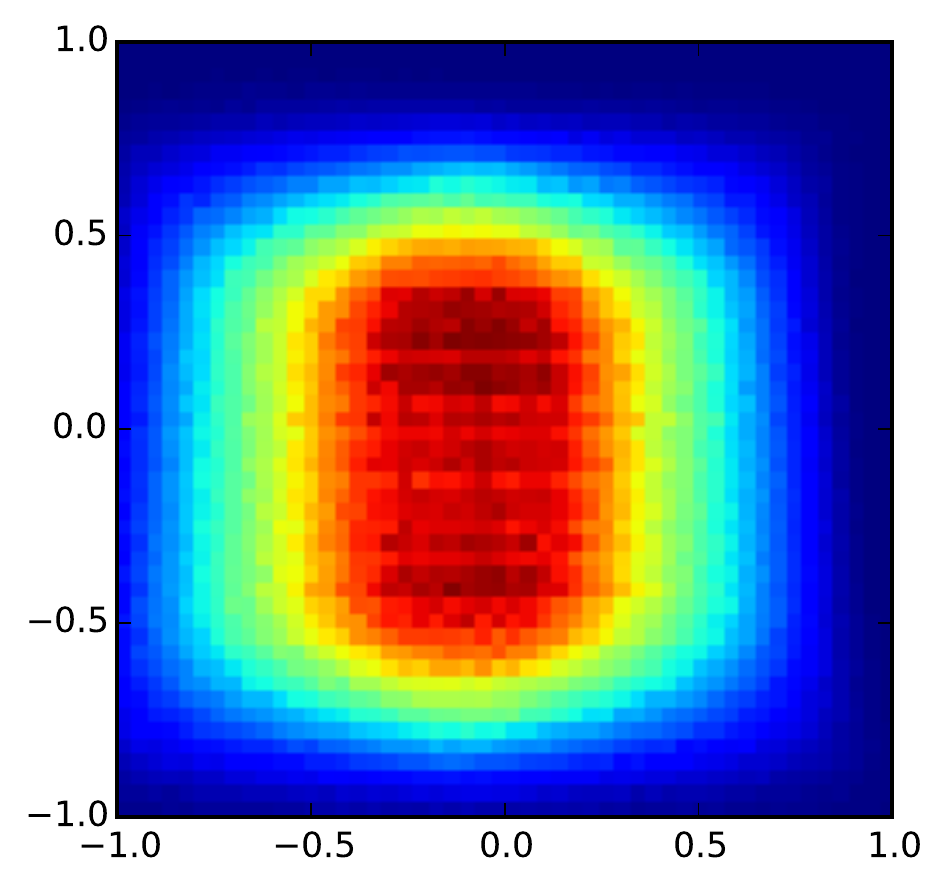}
                \caption{$\bm{g}$ (UT Multiview)}
                \label{fig:dataset_comparison:utfull-gaze}
        \end{subfigure}%
        \begin{subfigure}[b]{0.33\linewidth}
                \includegraphics[width=\textwidth]{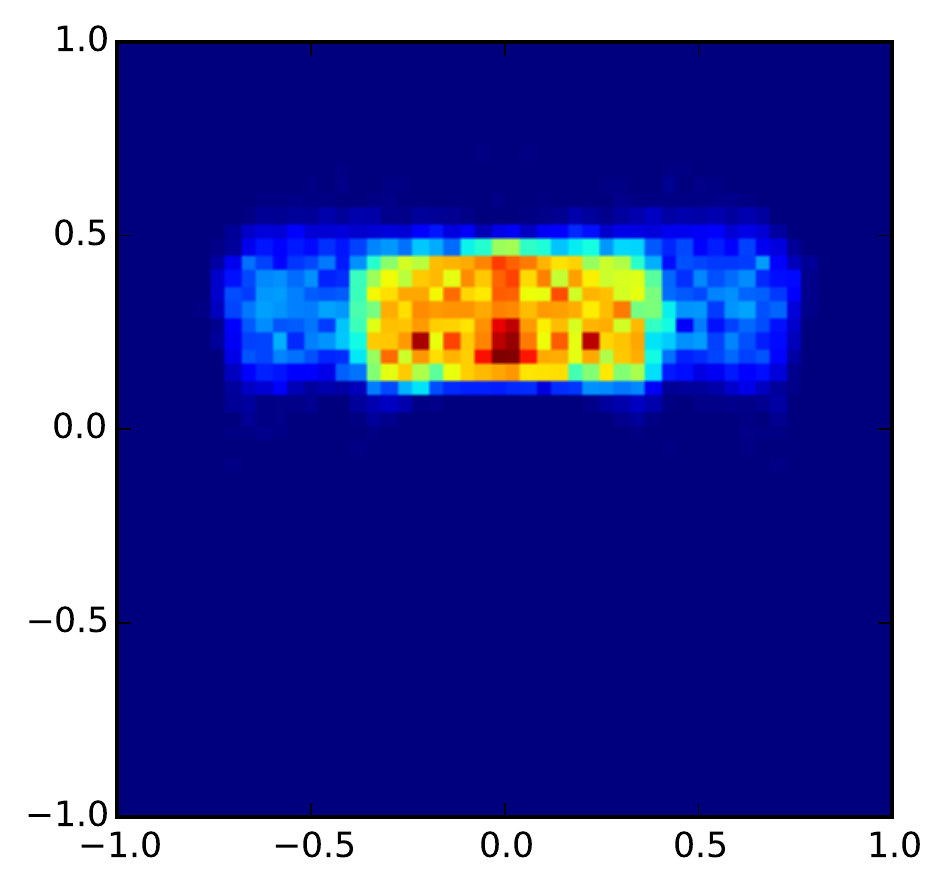}
                \caption{$\bm{g}$ (Eyediap)}
                \label{fig:dataset_comparison:idiap-ds-gaze}
        \end{subfigure}%
    \caption{Distributions of head angle ($\bm{h}$) and gaze angle ($\bm{g}$) for the \datasetname, UT Multiview, and Eyediap~datasets (cf. Table~\ref{tab:other_datasets}).}
    \label{fig:dataset_comparison}
\end{figure}

To further characterise our \datasetname~dataset in comparison with the other recent datasets~\cite{FunesMora_ETRA_2014,suganolearning}, Figure~\ref{fig:dataset_comparison} summarises distributions of the head and gaze angles $\bm{h}, \bm{g}$ in the normalised space. The normalisation was done as described in Section~\ref{subsec:normalisation}. Each figure corresponds to a 2D histogram of either pose or gaze, colour-coded from blue (minimum) to red (maximum). Although the UT Multiview dataset (see Figures~\ref{fig:dataset_comparison:utfull-pose} and~\ref{fig:dataset_comparison:utfull-gaze}) is recorded under a controlled lighting condition, it contains synthesised eye images which largely cover both gaze and pose spaces. Although the Eyediap dataset has mainly two different gaze targets, Figures~\ref{fig:dataset_comparison:idiap-ds-pose} and~\ref{fig:dataset_comparison:idiap-ds-gaze} show distributions of 2D screen targets, which is closer to our setting.
Our \datasetname~dataset covers a 2D screen space as in the Eyediap dataset; however, the gaze angle distributions are not overlapping, due to the difference in camera positions (see Figures~\ref{fig:dataset_comparison:mpii-pose} and~\ref{fig:dataset_comparison:mpii-gaze}).
This indicates that the Eyediap dataset does not cover the range of gaze directions that can occur during laptop interactions and that our \datasetname~can serve as a more appropriate basis for training and testing gaze estimators.

\begin{figure}[t]
  \centering
        \begin{subfigure}[b]{0.5\linewidth}
                \centering
                \includegraphics[width=0.9\textwidth]{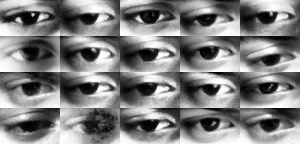}
                \caption{\datasetname}
                \label{fig:dataset_comparison:mpii-images}
        \end{subfigure}%
        \begin{subfigure}[b]{0.5\linewidth}
                \centering
                \includegraphics[width=0.9\textwidth]{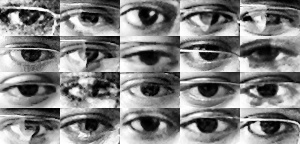}
                \caption{\datasetname}
                \label{fig:dataset_comparison:mpii-images-glasses}
        \end{subfigure}%

        \begin{subfigure}[b]{0.5\linewidth}
                \centering
                \includegraphics[width=0.9\textwidth]{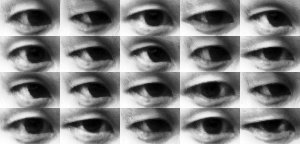}
                \caption{UT Multiview}
                \label{fig:dataset_comparison:ut-images}
        \end{subfigure}%
        \begin{subfigure}[b]{0.5\linewidth}
                \centering
                \includegraphics[width=0.9\textwidth]{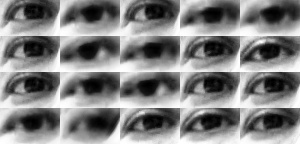}
                \caption{Eyediap}
                \label{fig:dataset_comparison:idiap-images}
        \end{subfigure}%

    \caption{Example images from the \datasetname~(non-eyeglasses and wearing eyeglasses), UT Multiview, Eyediap~datasets.}
    \label{fig:dataset_comparison_image}
\end{figure}

Figure~\ref{fig:dataset_comparison_image} shows sample eye images from each dataset after the normalisation (see Section~\ref{subsec:normalisation}).
Each group of images was randomly selected from a single person for roughly the same gaze directions.
Compared to the UT Multiview and Eyediap datasets (see Figures~\ref{fig:dataset_comparison:ut-images} and~\ref{fig:dataset_comparison:idiap-images}), our \datasetname~dataset contains larger appearance variations even inside the eye region (see Figure~\ref{fig:dataset_comparison:mpii-images}).
The variation becomes more significant in the case of a person wearing eyeglasses (see Figure~\ref{fig:dataset_comparison:mpii-images-glasses}), and they depict the complexity of the daily-life setting in terms of appearance changes.

\section{Method}
\label{sec:method}

Figure~\ref{fig:pipeline} provides an overview of our proposed method for in-the-wild appearance-based gaze estimation using multimodal convolutional neural networks (CNN). 
We first employ state-of-the-art face detection and facial landmark detection methods to locate landmarks in the input image obtained from the calibrated monocular RGB camera.
We then fit a generic 3D facial shape model to estimate 3D poses of the detected faces and apply the space normalisation technique proposed in~\cite{suganolearning} to crop and warp the head pose and eye images to the normalised training space.
The CNN is used to learn the mapping from the head poses and eye images to gaze directions in the camera coordinate system.

\subsection{Face Alignment and 3D Head Pose Estimation}

Our method first detects the user's face in the image using Li et al.'s SURF cascade method~\cite{Li2013SURF}.
We assume a single face in the images and take the largest bounding box if the detector returns multiple face proposals. We discard all images in which the detector fails to find any face, which happened in about 5\% of all cases.
Afterwards, we use Baltru\v{s}aitis et al.'s constrained local mode framework to detect facial landmarks~\cite{Baltrusaitis2014CCNF}.

We use the same definition of the face model and head coordinate system as~\cite{suganolearning}.
The face model consists of 3D positions of six facial landmarks (eye and mouth corners, cf. Figure~\ref{fig:pipeline}). 
The head coordinate system is defined according to the triangle connecting three midpoints of the eyes and mouth.
We fit the model by estimating the initial solution using the EPnP algorithm~\cite{Lepetit2009EPnP}, and further refining the pose via non-linear optimisation.
3D head rotation $\bm{r}$ is defined as the rotation from the head coordinate system to the camera coordinate system, and the eye position $\bm{t}$ is defined as the midpoint of eye corners for each eye.

While previous works assumed accurate head poses, we use a generic mean facial shape model for the 3D pose estimation to evaluate the whole gaze estimation pipeline in a practical setting.
3D positions of the six landmarks are recorded from all of the participants using an external stereo camera prior to the data collection, and the generic shape is built as the mean shape across all participants.

\subsection{Data Normalisation}
\label{subsec:normalisation}

Similar to~\cite{suganolearning}, we normalise the image and head pose space into a polar-coordinate angle space.
Fundamentally speaking, object pose has six degrees of freedom, and in the simplest case the gaze estimator has to handle eye appearance changes in this 6D space.
However, since arbitrary scaling and rotation of the camera can be compensated for by its corresponding perspective image warping, the appearance variation that needs to be handled inside the appearance-based estimation function has only two degrees of freedom.
The task of pose-independent appearance-based gaze estimation is to learn the mapping between gaze directions and eye appearances, which cannot be compensated for by virtually rotating and scaling the camera.

Briefly, the normalisation is done by scaling and rotating the camera so that: 1) the camera looks at the midpoint of the eye corners from a fixed distance $d$, and 2) $x$ axes of the head coordinate system and camera coordinate system become parallel.
Eye images were cropped at a fixed resolution $W \times H$ with a fixed focal length $f$ in the normalised camera space, and histogram-equalised to form the input eye image.
This results in a set of fixed-resolution eye images $\bm{e}$ and 2D head angle vectors $\bm{h}$, and the ground-truth gaze positions are also converted to the normalised camera space to give 2D gaze angle (yaw and pitch) vectors $\bm{g}$.
In order to reduce the effect of different lighting conditions, eye images $\bm{e}$ are histogram-equalised after the normalisation process.
We used the same setting for camera distance $d$, focal length $f$ and the resolution $W \times H$ as in~\cite{suganolearning}. In this manner, the normalised eye images are compatible between different datasets and we can evaluate the cross-dataset performance of appearance-based methods.

\subsection{Gaze Estimation With Multimodal CNNs}

The task for the CNN is to learn the mapping from the input features (2D head angle $\bm{h}$ and eye image $\bm{e}$) to gaze angles $\bm{g}$ in the normalised space.
As pointed out in \cite{suganolearning}, the difference between the left and right eyes is irrelevant in the person-independent training scenario. By flipping eye images horizontally and mirroring $\bm{h}$ and $\bm{g}$ around the $y$ axis, we handle both eyes by a single regression function.

\begin{figure}[t]
\center
\includegraphics[width=\columnwidth]{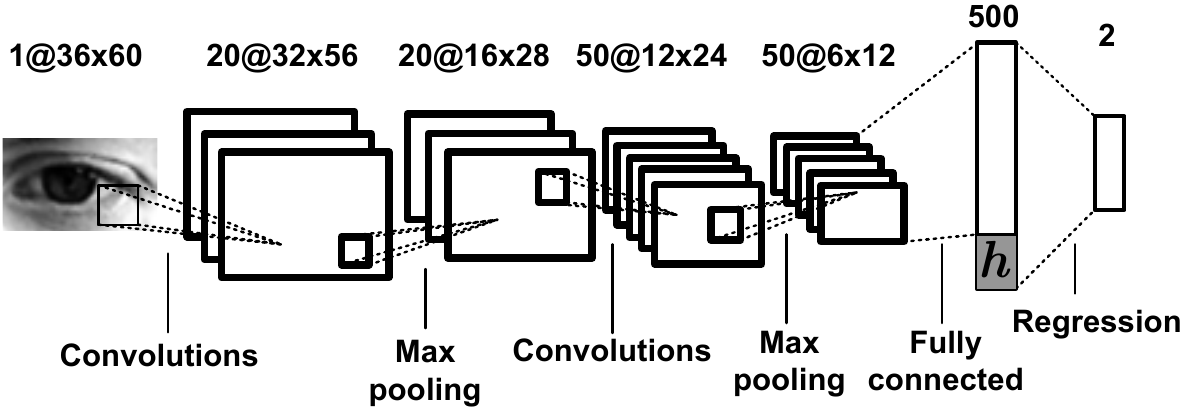}
\caption{Architecture of the proposed multimodal CNN. Head angle vectors $\bm{h}$ are added to the output of the fully connected layer.}
\label{fig:cnnmodel}
\end{figure}

Our model uses the LeNet network architecture that consists of one convolutional layer followed by a max-pooling layer, a second convolution layer followed by a max-pooling layer, and a final fully connected layer~\cite{lecun1998gradient,jia2014caffe}.
We train a linear regression layer on top of the fully connected layer to predict gaze angle vectors $\bm{g}$.
We use a multimodal CNN model to take advantage of both eye image and head pose information~\cite{ngiam2011multimodal}.
We encode head pose information into our CNN model by concatenating $\bm{h}$ with the output of the fully connected layer~(see Figure~\ref{fig:cnnmodel}).
Input to the network are the grey-scale eye images $\bm{e}$ with a fixed size of $60 \times 36$ pixels. For the two convolutional layers, the feature size is $5 \times 5$ pixels, while the number of features is 20 for the first layer and 50 for the second layer. The number of hidden units in the fully connected layer is 500, where each unit connects to all the feature maps of the previous convolutional layer, and is calculated by summing up all activation values. The output of the network is a 2D gaze angle vector $\bm{\hat g}$ that consists of two gaze angles, yaw $\hat g_{\phi}$ and pitch $\hat g_{\theta}$.
As a loss function we use the sum of the individual ${L_2}$ losses that measure the distance between the predicted $\bm{\hat g}$ and actual gaze angle vectors $\bm{g}$.

\section{Experiments}

In this section, we discuss the person-independent gaze estimation task and validate the effectiveness of the proposed CNN-based gaze estimation approach.
We conduct both cross-dataset and within-dataset conditions to compare our method with state-of-the-art methods on the \datasetname~dataset.
To account for the sample number bias among participants in our dataset, in the following experiments we use a random subset for both training and testing. Specifically, we randomly pick 1,500 left eye samples and 1,500 right eye samples from each person\footnote{Since one participant has only 1,448 images, we randomly oversampled the data to get 3,000.}.

In addition to our CNN-based method, we evaluate the following baseline methods using the same facial landmark detection, head pose estimation, and input features.

\vspace{-0.3cm}
\paragraph{Random Forests (RF)}
Random forests were recently demonstrated to outperform existing methods for person-independent appearance-based gaze estimation~\cite{suganolearning}. We use the implementation provided by the authors, which first clusters training samples according to head angles and query test samples to their nearest clusters. We used the same parameters as in~\cite{suganolearning}, and also resized input eye images to $16 \times 9$ pixels.

\vspace{-0.3cm}
\paragraph{$k$-Nearest Neighbours (kNN)}
As shown in~\cite{suganolearning}, a simple kNN regression estimator can perform well in scenarios that offer a large amount of dense training samples. We use the same kNN implementation and also incorporate a training sample clustering in head angle space.

\vspace{-0.3cm}
\paragraph{Adaptive Linear Regression (ALR)}
Because it was originally designed for a person-specific and sparse set of training samples~\cite{lu2014alr}, ALR does not scale to large datasets. We therefore use the same approximation as in~\cite{odobez2013person}, i.e.\ we select five training persons for each test person by evaluating the interpolation weights. We further select random subsets of samples from the test sample's neighbours in head pose space. We use the same image resolution as for RF.

\vspace{-0.3cm}
\paragraph{Support Vector Regression (SVR)}
Schneider et al.~\cite{schneider2014manifold} used SVR with a polynomial kernel under a fixed head pose. We use a linear SVR~\cite{fan2008liblinear} given the large amount of training data. We also use a concatenated vector of HOG and LBP features ($6 \times 4$ blocks, $2 \times 2$ cells for HOG) as suggested in~\cite{schneider2014manifold}. However, we do not use manifold alignment, since it does not support pose-independent training.

\vspace{-0.3cm}
\paragraph{Shape-Based Approach (EyeTab)}
Finally, in addition to these appearance-based methods, we evaluate one state-of-the-art shape-based method~\cite{wood14_etra} on the \datasetname~dataset. We use the implementation provided by the authors. In their method gaze estimation is performed by fitting a limbus model (a fixed-diameter disc) to detected iris edges.

\subsection{Cross-Dataset Evaluation}
\label{sec:cross_dataset}

We first present a comparative experimental validation for the cross-dataset evaluation condition.
We selected the UT Multiview dataset as the training dataset because it covers the largest area in head and gaze angle space (see Figure~\ref{fig:dataset_comparison}).
In addition to our \datasetname~dataset, we also show results using the Eyediap dataset as test data.

For the Eyediap dataset we used the 3D head poses provided with the dataset. These were estimated by fitting personal 3D shape models to depth images~\cite{FunesMora_ETRA_2014}.
Since their floating target sequences contain many extreme gaze directions that are not covered by the UT Multiview dataset, we only used the screen target sequences.

\begin{figure}[t]
\center
\includegraphics[width=\columnwidth]{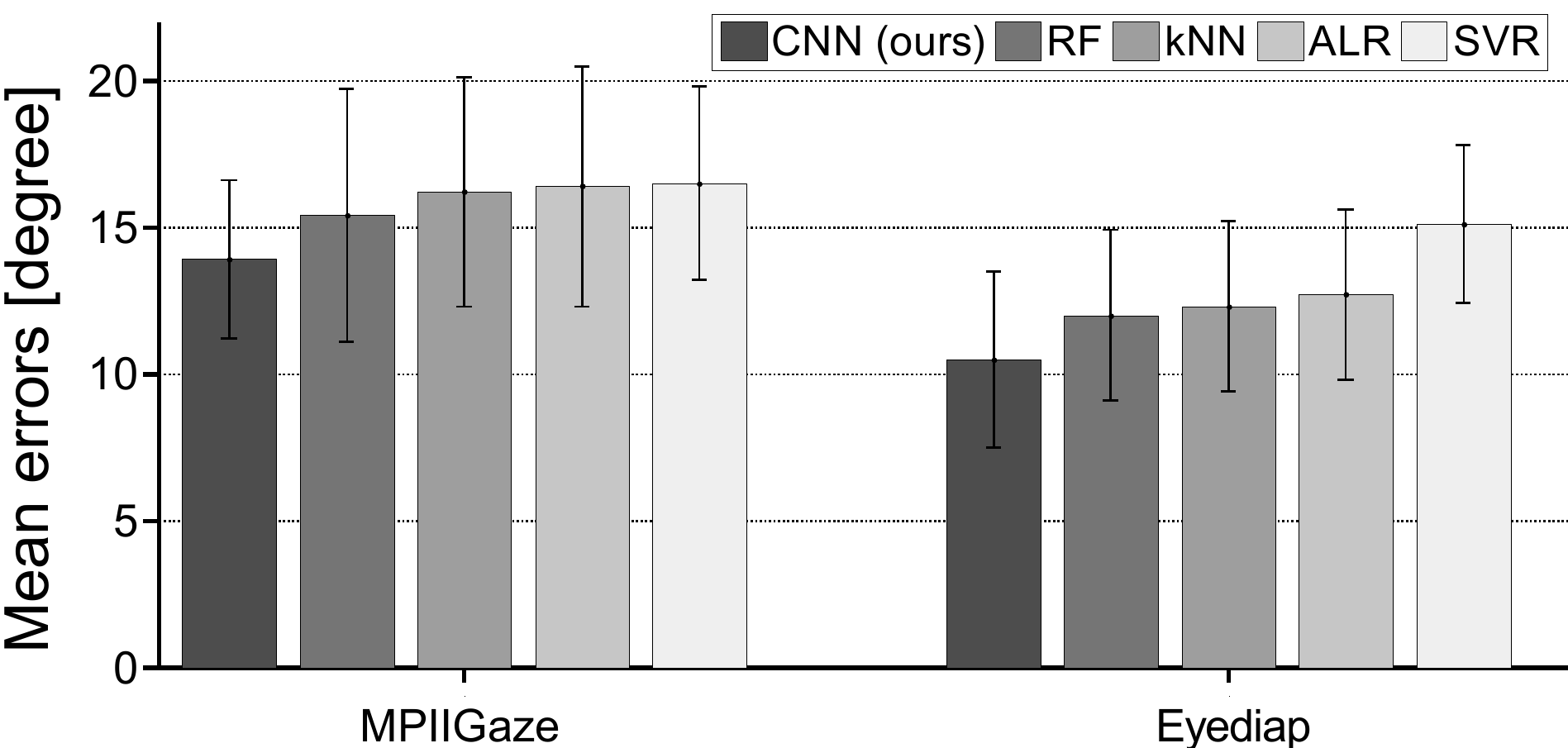}
\caption{Cross-dataset evaluation with training data from the UT Multiview dataset. Bars correspond to mean error across all participants in the \datasetname~(left) and screen-target sequences of Eyediap (right) datasets. Error bars indicate standard deviations.}
\label{fig:cross_dataset}
\end{figure}

Figure~\ref{fig:cross_dataset} summarises mean angular errors of all methods on both \datasetname~and Eyediap.
Bars correspond to mean error across all participants in each dataset, and error bars indicate standard deviations across persons.
The mean prediction error of a naive predictor that always outputs the average gaze direction of all training samples is 42.4 degrees on Eyediap and 34.2 degrees on \datasetname.
The shape-based EyeTab method performs poorly on the \datasetname~dataset (47.1 degrees mean error and 7\% misdetection rate), and this supports the advantage of the appearance-based approaches in challenging conditions.
In this setting, our CNN-based approach shows the best accuracy on both datasets (13.9 degrees on \datasetname, 10.5 degrees on Eyediap), with a significant performance gain (10\% on \datasetname, 12\% on Eyediap, paired Wilcoxon test \cite{wilcoxon1945individual}, p $<$ 0.05) over the state-of-the-art RF method.
However, performance on \datasetname~is generally worse than on the Eyediap dataset, which indicates the fundamental difficulty of the in-the-wild setting.

While our CNN-based approach expanded the feasibility of the generalisation task, these results at the same time reveal the critical limitation of the UT Multiview dataset and their learning-by-synthesis approach, whose variation of training data is limited in terms of eye appearances.
This indicates the importance of the training data, and that we need to address this goal from the both standpoints of data and methodology to bridge the gap from the domain-restricted training scenario.

\subsection{Within-Dataset Evaluation}\label{sec:within_dataset}

Although the previous cross-dataset evaluation showed the advantage of our CNN-based gaze estimation approach, there is still a huge performance gap compared to the performance reported in~\cite{suganolearning}.
To discuss the limits of person-independent performance on the \datasetname~dataset, we performed leave-one-person-out evaluation on the \datasetname~dataset.

\begin{figure}[t]
\center
\includegraphics[width=\columnwidth]{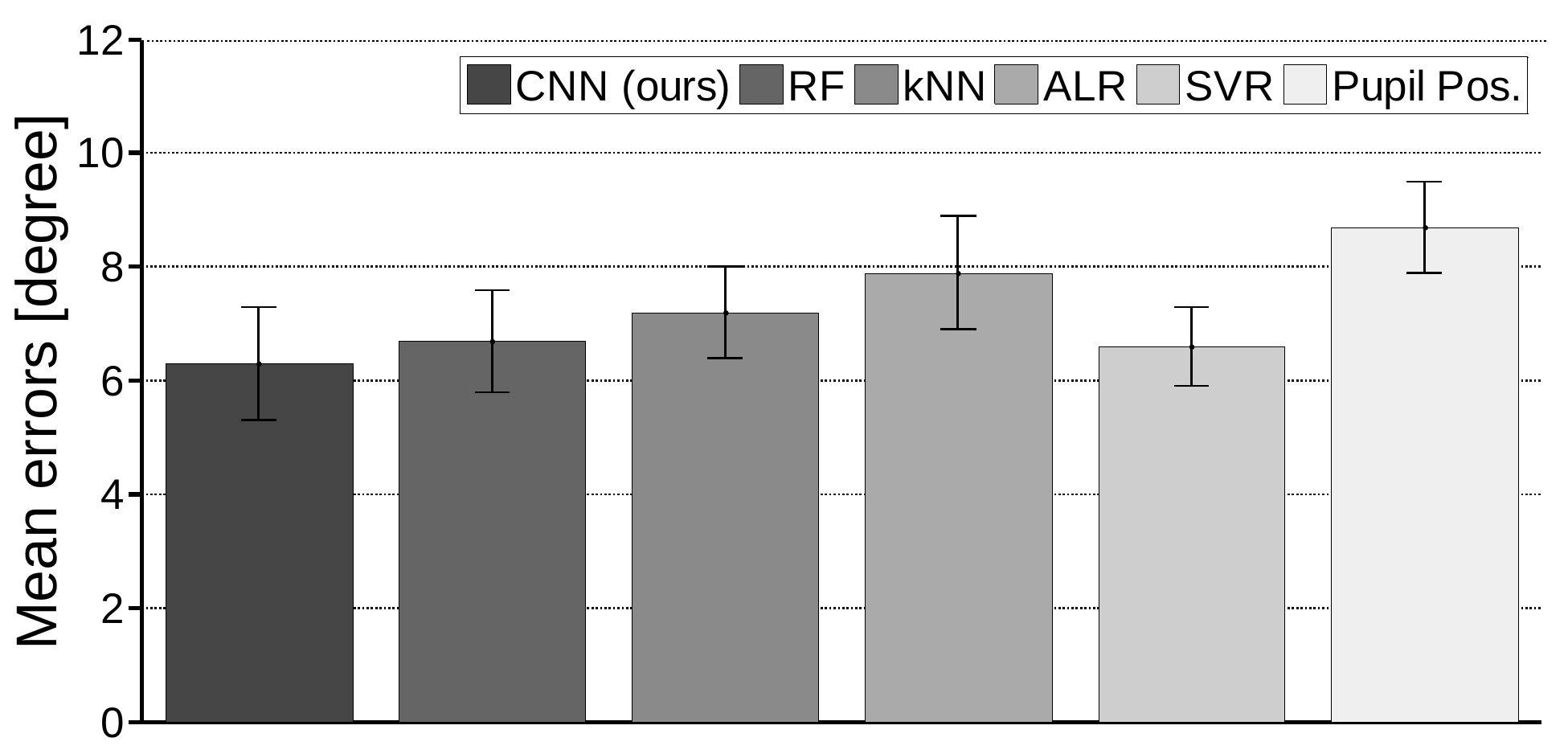}
\caption{Within-dataset leave-one-person-out evaluation on \datasetname. Mean estimation errors of the proposed method and other appearance-based methods. Error bars indicate standard deviations.}
\label{fig:within_dataset}
\end{figure}

With the same baseline methods as in Section~\ref{sec:cross_dataset}, Figure~\ref{fig:within_dataset} shows mean angular errors of the within-dataset evaluation.
Since the model-based EyeTab method has been shown to perform relatively poorly in our setting, we alternatively show a learning-based result using the detected pupil (iris centre) positions. More specifically, we used the pupil positions detected using~\cite{wood14_etra} in the normalised eye image space as a feature for kNN regression, and performed the same leave-one-person-out test.

In this case there is domain-specific prior knowledge about gaze distribution, and the mean prediction error becomes 13.9 degrees.
The pupil position-based approach works better than the original EyeTab method but its performance is still worse than appearance-based gaze estimation methods.
All appearance-based methods showed better performances than in Section~\ref{sec:cross_dataset}, and this indicates the importance of dataset- or domain-specific training data for appearance-based gaze estimation methods.
Although its performance gain over the other baseline methods becomes smaller in this setting, our CNN-based method still performed the best among them with 6.3 degrees mean error.

\begin{figure}[t]
\center
\includegraphics[width=\columnwidth]{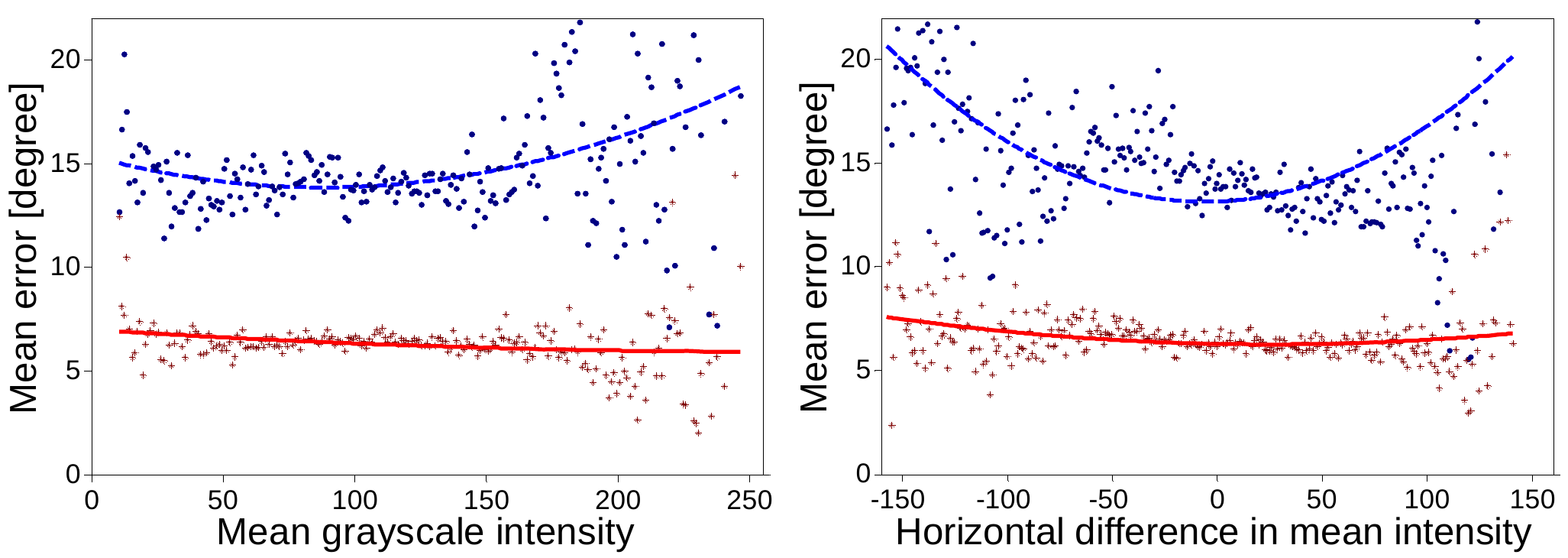}
\caption{Estimation error distribution w.r.t. mean grey-scale intensity within the face region (left) and horizontal mean grey-scale
intensity difference between the left and right half of the face region (right). The blue dots and curve from cross-dataset evaluation and the red dots and curve from within-\datasetname~evaluation.}
\label{fig:illumination}
\end{figure}

In order to illustrate the difference on handling appearance variations between cross-dataset and within-dataset scenarios, Figure~\ref{fig:illumination} shows estimation errors with respect to different illumination conditions.
Similarly to Figure~\ref{fig:characteristics}, we evaluate the error distribution with respect to mean grey-scale intensity of the face region and horizontal mean intensity difference between the left and right face regions.
Compared to the model pre-trained on the UT Multiview dataset (blue dots and curve), the model trained on the \datasetname~dataset (red dots and curve) shows better performance across different lighting conditions.
This clearly illustrates the effect of different lighting conditions and the importance of the appearance variation in the training data.

\subsection{Performance Validation of the Multimodal CNN}
\label{sec:within_ut}

While previous results show the potential of appearance-based gaze estimation methods in a challenging daily-life condition, there still exists a large performance gap compared to person-specific training results reported in prior work.
To further discuss the performance limits of the CNN-based approach, we also show more detailed comparisons between RF and CNN models.

\begin{figure}[t]
\center
\includegraphics[width=\columnwidth]{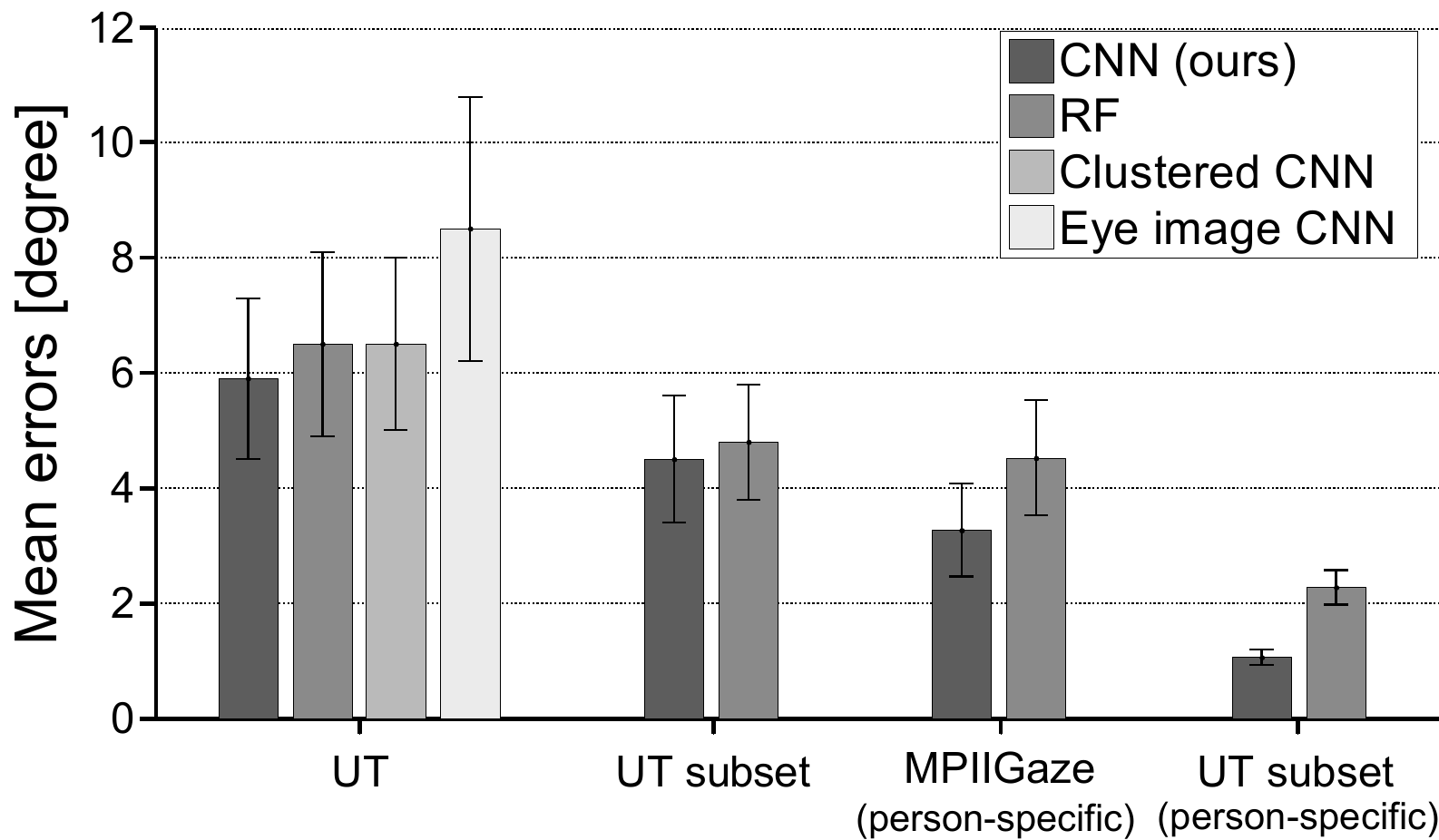}
\caption{Comparison of the different CNN models and RF on (from left to right): UT Multiview dataset, subset of the UT Multiview dataset which has the same head and gaze angle ranges as the \datasetname~dataset, using person-specific training on the \datasetname~dataset, and using person-specific training on the UT Multiview subset. Error bars indicate standard deviations.}
\label{fig:CNN_UT_UTsub}
\end{figure}

We first show a comparison between different architectures of the CNN on the UT Multiview dataset with the same three-fold cross-validation setting as reported in~\cite{suganolearning} (see Figure~\ref{fig:CNN_UT_UTsub} left).
As can be seen, our proposed multimodal CNN model outperformed the RF method with 5.9 degrees mean error.
Although \cite{suganolearning} reported that their pose-clustered structure improved their RF performance, the performance of the CNN became worse if the same clustering structure was introduced. This indicates the higher learning flexibility of the CNN, which contributes to the large performance gain in the cross-dataset case (Section~\ref{sec:cross_dataset}).
The performance can be degraded further if there is no head pose input, and this shows the fundamental importance of the head pose information in this pose-independent gaze estimation task.

The performance within the UT Multiview dataset is almost in the same range as the performance within the \datasetname~dataset  (see Figure~\ref{fig:within_dataset}). However, these two cases are expected to have different difficulty levels.
To investigate the difference within these results in more detail, we further show a three fold evaluation using a subset (3,000 samples per person) of the UT Multiview dataset selected so as to have the same pose and gaze angle distributions as the \datasetname~dataset.
The result is shown in the next part of Figure~\ref{fig:CNN_UT_UTsub}, and the performance gap compared to Figure~\ref{fig:within_dataset} indicates the error that arises from the in-the-wild setting, including appearance variations and eye alignment errors.

Although this is not always a practical assumption, estimators trained on person-specific training data show the upper limit of the performance we can achieve.
The rest of Figure~\ref{fig:CNN_UT_UTsub} shows mean errors of person-specific models on both the \datasetname~and UT Multiview datasets.
For \datasetname, the last quarter of the data from each person was used as test data, and the rest of the data was used as training data.
For UT Multiview, 500 test samples were randomly selected for each person from the above subset, and the other 2,500 samples were used as training data.
These results further show the potential performance of the appearance-based estimator, and clearly depict the performance gap to be investigated.

\section{Conclusion}

Despite a large body of previous work on the topic, appearance-based gaze estimation methods have so far been evaluated exclusively under controlled laboratory conditions.
In this work, we presented the first extensive study on appearance-based gaze estimation in the unconstrained daily-life setting.
We built a novel in-the-wild gaze dataset through a long-term data collection using laptops, which shows significantly larger variations in eye appearance than existing datasets.
Throughout the comprehensive benchmarking of image-based monocular gaze estimation methods, our study clearly revealed the potential and remaining technical challenges of appearance-based gaze estimation.
Our CNN-based estimation model significantly outperforms state-of-the-art methods in the most challenging person- and pose-independent training scenario.
This work and our dataset provide a critical insight on addressing grand challenges in daily-life gaze interaction.

\section*{Acknowledgements}

We would like to thank Laura Sesma for her help with the dataset handling and normalisation. This work was funded in part by the Cluster of Excellence on Multimodal Computing and Interaction (MMCI) at Saarland University as well as an Alexander von Humboldt Research Fellowship. We would further like to thank the NVIDIA Corporation for donating the GPU used in this research.

{\small
\bibliographystyle{ieee}
\bibliography{references}

\begin{thebibliography}{10}\itemsep=-1pt

\bibitem{alnajarcalibration}
F.~Alnajar, T.~Gevers, R.~Valenti, and S.~Ghebreab.
\newblock Calibration-free gaze estimation using human gaze patterns.
\newblock In {\em Proc. ICCV}, 2013.

\bibitem{Baltrusaitis2014CCNF}
T.~Baltru\v{s}aitis, P.~Robinson, and L.-P. Morency.
\newblock Continuous conditional neural fields for structured regression.
\newblock In {\em Proc. ECCV}, pages 593--608, 2014.

\bibitem{baluja1994non}
S.~Baluja and D.~Pomerleau.
\newblock Non-intrusive gaze tracking using artificial neural networks.
\newblock Technical report, DTIC Document, 1994.

\bibitem{Chen2008}
J.~Chen and Q.~Ji.
\newblock 3d gaze estimation with a single camera without ir illumination.
\newblock In {\em Proc. ICPR}, pages 1--4, 2008.

\bibitem{chen2011probabilistic}
J.~Chen and Q.~Ji.
\newblock Probabilistic gaze estimation without active personal calibration.
\newblock In {\em Proc. CVPR}, pages 609--616, 2011.

\bibitem{choi2013appearance}
J.~Choi, B.~Ahn, J.~Parl, and I.~S. Kweon.
\newblock Appearance-based gaze estimation using kinect.
\newblock In {\em Proc. URAI}, pages 260--261, 2013.

\bibitem{fan2008liblinear}
R.-E. Fan, K.-W. Chang, C.-J. Hsieh, X.-R. Wang, and C.-J. Lin.
\newblock Liblinear: A library for large linear classification.
\newblock {\em The Journal of Machine Learning Research}, 9:1871--1874, 2008.

\bibitem{FunesMora_ETRA_2014}
K.~A. Funes~Mora, F.~Monay, and J.-M. Odobez.
\newblock Eyediap: A database for the development and evaluation of gaze
  estimation algorithms from rgb and rgb-d cameras.
\newblock In {\em Proc. ETRA}, pages 255--258, 2014.

\bibitem{funes2012gaze}
K.~A. Funes~Mora and J.-M. Odobez.
\newblock Gaze estimation from multimodal kinect data.
\newblock In {\em Proc. CVPRW}, pages 25--30, 2012.

\bibitem{odobez2013person}
K.~A. Funes~Mora and J.-M. Odobez.
\newblock Person independent 3d gaze estimation from remote rgb-d cameras.
\newblock In {\em Proc. ICIP}, 2013.

\bibitem{mora2014geometric}
K.~A. Funes~Mora and J.-M. Odobez.
\newblock Geometric generative gaze estimation (g3e) for remote rgb-d cameras.
\newblock In {\em Proc. CVPR}, pages 1773--1780, 2014.

\bibitem{hansen2010eye}
D.~W. Hansen and Q.~Ji.
\newblock In the eye of the beholder: A survey of models for eyes and gaze.
\newblock {\em IEEE Transactions on Pattern Analysis and Machine Intelligence},
  32(3):478--500, 2010.

\bibitem{hennessey2006single}
C.~Hennessey, B.~Noureddin, and P.~Lawrence.
\newblock A single camera eye-gaze tracking system with free head motion.
\newblock In {\em Proc. ETRA}, pages 87--94, 2006.

\bibitem{huang2008labeled}
G.~B. Huang, M.~Mattar, T.~Berg, E.~Learned-Miller, et~al.
\newblock Labeled faces in the wild: A database forstudying face recognition in
  unconstrained environments.
\newblock In {\em Workshop on Faces in'Real-Life'Images: Detection, Alignment,
  and Recognition}, 2008.

\bibitem{Huang:2014:BSE:2647868.2655031}
M.~X. Huang, T.~C. Kwok, G.~Ngai, H.~V. Leong, and S.~C. Chan.
\newblock Building a self-learning eye gaze model from user interaction data.
\newblock In {\em Proc. MM}, pages 1017--1020, 2014.

\bibitem{Ishikawa_2004_4705}
T.~Ishikawa, S.~Baker, I.~Matthews, and T.~Kanade.
\newblock Passive driver gaze tracking with active appearance models.
\newblock In {\em Proc. 11th World Congress on Intelligent Transportation
  Systems}, 2004.

\bibitem{jia2014caffe}
Y.~Jia, E.~Shelhamer, J.~Donahue, S.~Karayev, J.~Long, R.~Girshick,
  S.~Guadarrama, and T.~Darrell.
\newblock Caffe: Convolutional architecture for fast feature embedding.
\newblock {\em arXiv preprint arXiv:1408.5093}, 2014.

\bibitem{jianfeng2014eye}
L.~Jianfeng and L.~Shigang.
\newblock Eye-model-based gaze estimation by rgb-d camera.
\newblock In {\em Proc. CVPRW}, pages 606--610, 2014.

\bibitem{larson1983experience}
R.~Larson and M.~Csikszentmihalyi.
\newblock The experience sampling method.
\newblock {\em New Directions for Methodology of Social \& Behavioral Science},
  1983.

\bibitem{lecun1998gradient}
Y.~LeCun, L.~Bottou, Y.~Bengio, and P.~Haffner.
\newblock Gradient-based learning applied to document recognition.
\newblock {\em Proceedings of the IEEE}, 86(11):2278--2324, 1998.

\bibitem{Lepetit2009EPnP}
V.~Lepetit, F.~Moreno-Noguer, and P.~Fua.
\newblock {EPnP}: An accurate o(n) solution to the {PnP} problem.
\newblock {\em International Journal of Computer Vision}, 81(2):155--166, 2009.

\bibitem{Li2013SURF}
J.~Li and Y.~Zhang.
\newblock Learning surf cascade for fast and accurate object detection.
\newblock In {\em Proc. CVPR}, pages 3468--3475, 2013.

\bibitem{li2014secrets}
Y.~Li, X.~Hou, C.~Koch, J.~Rehg, and A.~Yuille.
\newblock The secrets of salient object segmentation.
\newblock In {\em Proc. CVPR}, 2014.

\bibitem{Liang:2013:AGT:2509315.2509318}
K.~Liang, Y.~Chahir, M.~Molina, C.~Tijus, and F.~Jouen.
\newblock Appearance-based gaze tracking with spectral clustering and
  semi-supervised gaussian process regression.
\newblock In {\em Proc. ETSA}, pages 17--23, 2013.

\bibitem{lu2014learning}
F.~Lu, T.~Okabe, Y.~Sugano, and Y.~Sato.
\newblock Learning gaze biases with head motion for head pose-free gaze
  estimation.
\newblock {\em Image and Vision Computing}, 32(3):169 -- 179, 2014.

\bibitem{lu2012head}
F.~Lu, Y.~Sugano, T.~Okabe, and Y.~Sato.
\newblock Head pose-free appearance-based gaze sensing via eye image synthesis.
\newblock In {\em Proc. ICPR}, pages 1008--1011, 2012.

\bibitem{lu2014alr}
F.~Lu, Y.~Sugano, T.~Okabe, and Y.~Sato.
\newblock Adaptive linear regression for appearance-based gaze estimation.
\newblock {\em IEEE Trans. PAMI}, 36(10):2033--2046, Oct 2014.

\bibitem{majaranta2014eye}
P.~Majaranta and A.~Bulling.
\newblock Eye tracking and eye-based human--computer interaction.
\newblock In {\em Advances in Physiological Computing}, pages 39--65. Springer,
  2014.

\bibitem{McMurrough:2012:ETD:2168556.2168622}
C.~D. McMurrough, V.~Metsis, J.~Rich, and F.~Makedon.
\newblock An eye tracking dataset for point of gaze detection.
\newblock In {\em Proc. ETRA}, pages 305--308, 2012.

\bibitem{morimoto2002detecting}
C.~H. Morimoto, A.~Amir, and M.~Flickner.
\newblock Detecting eye position and gaze from a single camera and 2 light
  sources.
\newblock In {\em Proc. ICPR}, pages 314--317, 2002.

\bibitem{morimoto2005eye}
C.~H. Morimoto and M.~R. Mimica.
\newblock Eye gaze tracking techniques for interactive applications.
\newblock {\em Computer Vision and Image Understanding}, 98(1):4--24, 2005.

\bibitem{ngiam2011multimodal}
J.~Ngiam, A.~Khosla, M.~Kim, J.~Nam, H.~Lee, and A.~Ng.
\newblock Multimodal deep learning.
\newblock In {\em Proc. ICML}, pages 689--696, 2011.

\bibitem{Rodrigues2010Mirror}
R.~Rodrigues, J.~a. Barreto, and U.~Nunes.
\newblock Camera pose estimation using images of planar mirror reflections.
\newblock In {\em Proc. ECCV}, pages 382--395, 2010.

\bibitem{schneider2014manifold}
T.~Schneider, B.~Schauerte, and R.~Stiefelhagen.
\newblock Manifold alignment for person independent appearance-based gaze
  estimation.
\newblock In {\em Proc. ICPR}, 2014.

\bibitem{sewell10_chi}
W.~Sewell and O.~Komogortsev.
\newblock Real-time eye gaze tracking with an unmodified commodity webcam
  employing a neural network.
\newblock In {\em Ext. Abstracts CHI}, pages 3739--3744, 2010.

\bibitem{shih2004novel}
S.-W. Shih and J.~Liu.
\newblock A novel approach to 3-d gaze tracking using stereo cameras.
\newblock {\em IEEE Transactions on Systems, Man, and Cybernetics, Part B:
  Cybernetics}, 34(1):234--245, 2004.

\bibitem{smith2013gaze}
B.~A. Smith, Q.~Yin, S.~K. Feiner, and S.~K. Nayar.
\newblock Gaze locking: passive eye contact detection for human-object
  interaction.
\newblock In {\em Proc. UIST}, pages 271--280, 2013.

\bibitem{sugano2013appearance}
Y.~Sugano, Y.~Matsushita, and Y.~Sato.
\newblock Appearance-based gaze estimation using visual saliency.
\newblock {\em IEEE Trans. on PAMI}, 35(2):329--341, Feb 2013.

\bibitem{suganolearning}
Y.~Sugano, Y.~Matsushita, and Y.~Sato.
\newblock Learning-by-synthesis for appearance-based 3d gaze estimation.
\newblock In {\em Proc. CVPR}, pages 1821--1828, 2014.

\bibitem{sugano2008incremental}
Y.~Sugano, Y.~Matsushita, Y.~Sato, and H.~Koike.
\newblock An incremental learning method for unconstrained gaze estimation.
\newblock In {\em Proc. ECCV}, pages 656--667, 2008.

\bibitem{sun2014realtime}
L.~Sun, M.~Song, Z.~Liu, and M.-T. Sun.
\newblock Realtime gaze estimation with online calibration.
\newblock In {\em Proc. ICME}, pages 1--6, 2014.

\bibitem{tan2002appearance}
K.-H. Tan, D.~J. Kriegman, and N.~Ahuja.
\newblock Appearance-based eye gaze estimation.
\newblock In {\em Proc. WACV}, pages 191--195, 2002.

\bibitem{torralba2011unbiased}
A.~Torralba and A.~A. Efros.
\newblock Unbiased look at dataset bias.
\newblock In {\em Proc. CVPR}, pages 1521--1528. IEEE, 2011.

\bibitem{valenti2012combining}
R.~Valenti, N.~Sebe, and T.~Gevers.
\newblock Combining head pose and eye location information for gaze estimation.
\newblock {\em IEEE Transactions on Image Processing}, 21(2):802--815, 2012.

\bibitem{villanueva2013hybrid}
A.~Villanueva, V.~Ponz, L.~Sesma-Sanchez, M.~Ariz, S.~Porta, and R.~Cabeza.
\newblock Hybrid method based on topography for robust detection of iris center
  and eye corners.
\newblock {\em ACM Transactions on Multimedia Computing, Communications, and
  Applications}, 9(4):25, 2013.

\bibitem{weidenbacher07}
U.~Weidenbacher, G.~Layher, P.-M. Strauss, and H.~Neumann.
\newblock A comprehensive head pose and gaze database.
\newblock In {\em Proc. IET}, pages 455--458, 2007.

\bibitem{wilcoxon1945individual}
F.~Wilcoxon.
\newblock Individual comparisons by ranking methods.
\newblock {\em Biometrics bulletin}, pages 80--83, 1945.

\bibitem{williams2006sparse}
O.~Williams, A.~Blake, and R.~Cipolla.
\newblock Sparse and semi-supervised visual mapping with the {S}\^{} 3{GP}.
\newblock In {\em Proc. CVPR}, pages 230--237, 2006.

\bibitem{wood14_etra}
E.~Wood and A.~Bulling.
\newblock Eyetab: Model-based gaze estimation on unmodified tablet computers.
\newblock In {\em Proc. ETRA}, pages 207--210, 2014.

\bibitem{yamazoe2008remote}
H.~Yamazoe, A.~Utsumi, T.~Yonezawa, and S.~Abe.
\newblock Remote gaze estimation with a single camera based on facial-feature
  tracking without special calibration actions.
\newblock In {\em Proc. ETRA}, pages 245--250, 2008.

\bibitem{yoo2005novel}
D.~H. Yoo and M.~J. Chung.
\newblock A novel non-intrusive eye gaze estimation using cross-ratio under
  large head motion.
\newblock {\em Computer Vision and Image Understanding}, 98(1):25--51, 2005.

\bibitem{zhu2005eye}
Z.~Zhu and Q.~Ji.
\newblock Eye gaze tracking under natural head movements.
\newblock In {\em Proc. CVPR}, pages 918--923, 2005.

\bibitem{zhu2006nonlinear}
Z.~Zhu, Q.~Ji, and K.~P. Bennett.
\newblock Nonlinear eye gaze mapping function estimation via support vector
  regression.
\newblock In {\em Proc. ICPR}, pages 1132--1135, 2006.

\end{thebibliography}
}

\end{document}